\documentclass{article}

 \usepackage[main,final]{neurips_2026}

\usepackage[utf8]{inputenc} 
\usepackage[T1]{fontenc}    
\usepackage{hyperref}       
\usepackage{url}            
\usepackage{booktabs}       
\usepackage{amsfonts}       
\usepackage{nicefrac}       
\usepackage{microtype}      
\usepackage{xcolor}         
\usepackage{graphicx}
\usepackage{booktabs}

\usepackage{booktabs}
\usepackage{algorithm}
\usepackage{algpseudocode}
\usepackage{makecell}
\usepackage{multirow}

\usepackage{enumitem}
\usepackage{amsmath}
\usepackage{tikz}
\usepackage{amssymb}
\usepackage{tcolorbox}
\usepackage{float}

\usepackage{tabularx}

\title{BlenderFORGE: Framework for Optimizing Reactive 3D-Graphics Editing Ability of MLLMs}

\author{%
  \textbf{Zilin Guo$^{1}$\thanks{Equal contribution.} \quad Tianrui ZHANG$^{1}$\footnotemark[1] \quad Yi RONG$^{1}$\footnotemark[1]} \\
  \textbf{Yichen Liu$^{2}$ \quad Yuxin Guo$^{2}$ \quad Jingcheng Ni$^{2}$} \\
  \textbf{Lewei Lu$^{2}$ \quad Zehuan Wu$^{2}$ \quad Dan Xu$^{1}$} \\[3pt]
  {\normalfont $^{1}$The Hong Kong University of Science and Technology} \\
  {\normalfont $^{2}$SenseTime Research} \\
  {\normalfont\footnotesize\texttt{\{zguobd,tzhangbu,yrongad\}@connect.ust.hk}, \texttt{danxu@cse.ust.hk}} \\
  {\normalfont\footnotesize\texttt{\{liuyichen,guoyuxin,nijingcheng,luotto,wuzehuan\}@sensetime.com}}
  \\
  Project Page: \url{https://sensetime-fvg.github.io/BlenderFORGE/}
}

\begin{document}

\setcitestyle{square,numbers,comma}
\bibliographystyle{unsrtnat}
\renewcommand{\bibsection}{\section*{References}}

\maketitle

\begin{abstract}

    Multimodal Large Language Models (MLLMs) have shown promise as interactive agents, yet precise, programmatic 3D graphics editing remains difficult to train because existing systems largely rely on closed-source model APIs and lack deterministic, state-verifiable execution environments. We present BlenderFORGE, a trainable data-environment-optimization framework for open-weight MLLM agents that edit explicit Blender scenes through Blender Python (bpy) code generation. BlenderFORGE targets four fundamental state-verifiable editing task families: object placement, shape-key editing, lighting adjustment, and material modification. It integrates three components: (1) a scalable perturbation-based data generation pipeline built from curated assets and large-scale 3D scene repositories; (2) an isolated Blender Sandbox for deterministic code execution, structured state export, and closed-loop reward computation; and (3) a two-stage training pipeline that uses offline teacher trajectories for cold-start Supervised Fine-Tuning (SFT) and sandbox-computed rewards for multi-turn tool-agentic Reinforcement Learning (AgentLoop-RL). Experiments across the four task families show that the resulting open-weight Qwen3-VL-8B and Qwen3.5-9B BlenderFORGE agents substantially improve over their base models and achieve competitive performance with several zero-shot proprietary MLLM baselines on selected metrics. Additional cross-benchmark evaluations and camera-reframing experiments provide evidence of transfer beyond the original training setting.
\end{abstract}

\section{Introduction}
\label{sec:intro}
\begin{figure}[htbp]
    \centering
    \includegraphics[width=1.0\textwidth]{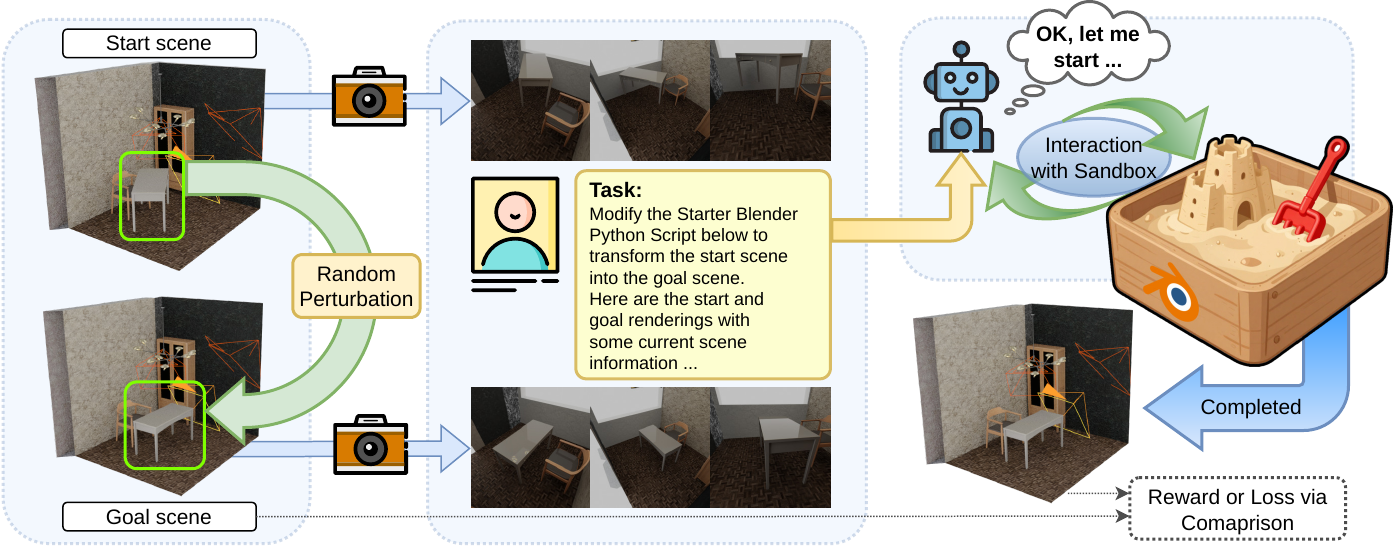}
    \caption{\textbf{Overview of the learning process under the BlenderFORGE framework.} MLLMs learn the correspondence between visual changes and 3D-graphics edits from scene perturbations.}
    \label{fig:teaser}
\end{figure}

The creation and manipulation of 3D content is central to a wide range of industries, from film production and game development to architectural visualization and robotics simulation.
Professional digital content creation (DCC) tools such as Blender~\cite{blender} provide powerful, fine-grained editing capabilities, yet demand substantial domain expertise and manual effort.
The recent emergence of Multimodal Large Language Models (MLLMs) offers a compelling pathway toward democratizing 3D graphics editing: by accepting natural language instructions alongside visual observations, an MLLM agent could potentially translate high-level user intent into precise, executable editing operations.
As illustrated in Fig.~\ref{fig:teaser}, our goal is to enable MLLMs to perceive a rendered 3D scene, reason about the required modifications, and generate Blender Python (\texttt{bpy}) code that achieves the desired edit.

Despite encouraging progress, current MLLM-based 3D editing systems are largely organized as inference-time search pipelines~\cite{Gu_2025_CVPR,huang2024blenderalchemyediting3dgraphics,hu2024scenecraft}. They generate candidate Blender edits and use rendered feedback or a verifier to select and refine edits over multiple rounds. This test-time scaling paradigm can improve individual task outcomes, but its gains are paid at deployment time and remain coupled to candidate diversity, verifier reliability, and proprietary model behavior. We instead study a complementary training-time direction that amortizes editing experience into an open-weight agent through parameter updates. This direction has not been explored for 3D graphics editing because training requires both scalable supervision from start-to-goal editing tasks and a high-throughput closed-loop environment for executable interaction, progress verification, and rollout collection.

To realize this training interface, we propose \textbf{BlenderFORGE}, a trainable closed-loop framework for open-weight MLLM agents that edit explicit Blender scenes through Blender Python (\texttt{bpy}) code. Across four 3D editing families, BlenderFORGE unifies task generation, code execution, verification, and optimization within a shared scene-state space. A scalable perturbation-based data pipeline samples start and goal states for object placement, shape-key editing, lighting adjustment, and material modification. A parallel, instance-isolated Blender environment executes agent-generated code, returns rendered observations, and exports structured scene states for deterministic verification. These components support a two-stage training pipeline that initializes the agent with offline teacher trajectories and refines it through AgentLoop-RL using sandbox-computed rewards.

In summary, our main contributions are:
\begin{enumerate}[leftmargin=*]
    \item \textbf{Scalable perturbation-based data generation.}
    We generate start-to-goal Blender editing tasks across placement, shape-key, lighting, and material edits, providing supervision and verifiable scene-state, visual, and geometric signals for SFT and RL.

    \item \textbf{A closed-loop environment for scalable agent optimization.}
    We support high-throughput rollouts with parallel, instance-isolated Blender execution, including \texttt{bpy} code execution, rendered feedback, structured scene-state export, and sandbox reward computation.

    \item \textbf{Training and analysis of open-weight editing agents.}
    We train Qwen3-VL-8B and Qwen3.5-9B with offline trajectory SFT and AgentLoop-RL using sandbox-computed rewards, showing consistent gains over their base models, competitive performance against several zero-shot or open-weight baselines on selected metrics, and ablations of each training stage.
\end{enumerate}

\section{Related works}
\label{sec:related_works}

\subsection{Representation-Level and Explicit 3D Editing}
A large body of 3D editing work operates on learned or reconstructed representations, including NeRFs, voxel fields, meshes, and 3D Gaussian splats~\cite{poole2022dreamfusion,kosiorek2021nerf,kerbl20233d,Chen_2024_CVPR,Hu_2025_CVPR,alliegro2023polydiff,siddiqui2024meshgpt,wang2022clip}. These methods are effective for representation-level generation or appearance editing, but they do not directly target executable manipulation of explicit DCC scene graphs. BlenderFORGE instead studies editing of existing Blender scenes through \texttt{bpy} actions, where success can be assessed through rendered views, exported scene state, and geometry.

\subsection{MLLM Agents for 3D Graphics}
Recent systems use LLMs or MLLMs to bridge language instructions and 3D software. SceneCraft~\cite{hu2024scenecraft}, LayoutGPT~\cite{feng2023layoutgpt}, LLM4CAD~\cite{li2024llm4cad}, and related systems~\cite{liu2025agentic,liu2025worldcraft,zhou2025scenex,yang2025sceneweaver} formulate scene synthesis or editing as code generation, planning, or tool use. BlenderAlchemy~\cite{huang2024blenderalchemyediting3dgraphics} shows that VLMs can edit Blender scenes through iterative visual feedback, while BlenderGym~\cite{Gu_2025_CVPR} provides a benchmark and studies inference-time scaling for graphics-editing foundation model systems. Building on BlenderGym's four task families and perceptual metrics, BlenderFORGE focuses on the complementary training-time problem: constructing data, environment feedback, and rewards for optimizing an open-weight MLLM editing agent.

\subsection{Interactive Software Agents and Training-Time Optimization}
Recent GUI and computer-use agents increasingly train visual-action models from GUI grounding data, action traces, or large-scale interaction corpora~\cite{hong2024cogagent,cheng2024seeclick,wu2024atlas,xu2024aguvis}. Web agents have also explored online and end-to-end multi-turn reinforcement learning to improve interactive decision-making beyond supervised imitation~\cite{qi2024webrl,wei2025webagent}. These works suggest that software-control agents can benefit from amortizing interaction experience into model parameters rather than relying only on prompting or test-time refinement. By contrast, executable 3D graphics editing lacks a comparable training substrate, with limited start-to-goal supervision and few stable environments for interactive SFT and RL. BlenderFORGE brings this training-time perspective to explicit 3D editing by turning Blender interaction into a scalable source of trajectories and verifiable rewards for MLLM optimization.

\section{Method}
\label{sec:method}
\begin{figure}[htbp]
    \centering
    \includegraphics[width=1.0\textwidth]{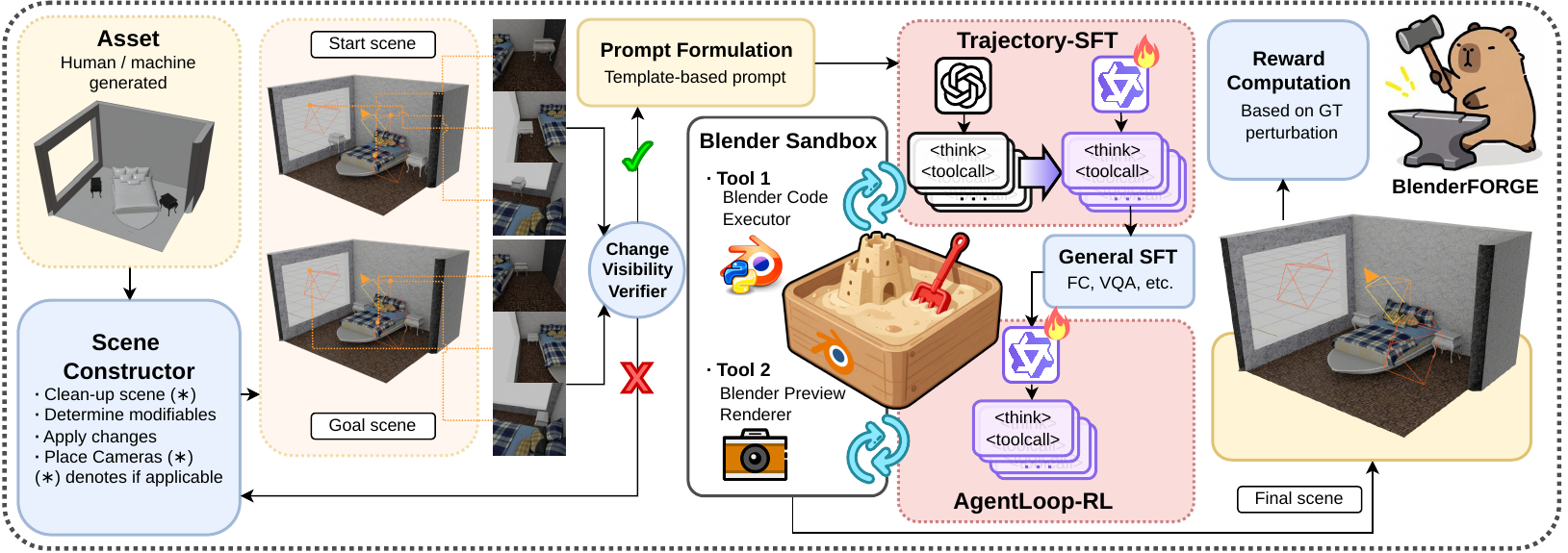}
    \caption{\textbf{The BlenderFORGE architecture}. BlenderFORGE is a unified framework that integrates a scalable 3D asset-based data generation pipeline with an interactive Blender sandbox to enable training MLLM agents in 3D graphics editing. (Left) Data Generation Pipeline: 3D assets are perturbed and filtered through a visibility check to ensure that the changes are observable. (Right) Training Loop: The validated data flows into training with an interactive Blender sandbox, enabling SFT and RL for MLLM agents.}
    \label{fig:main_graph}
\end{figure}

\subsection{Task Definitions}
\label{sec:task_definition}

We formulate executable 3D editing as a start-to-goal scene transformation problem. Each task specifies a start state $s_s$, a goal state $s_g$, a language instruction, and rendered observations available to the agent. Starting from $x_0=s_s$, the agent interacts with Blender over multiple steps by generating executable actions $a_k$ in Blender Python (\texttt{bpy}). Each action induces a scene transition $x_{k+1}=f(x_k,a_k)$, after which the agent may request rendered feedback $I(x_{k+1})$. Let $\mathcal{T}(u,v)$ denote the relative transformation from state $u$ to state $v$. The objective is
\begin{equation}
    \underset{\{a_k\}_{k=0}^{K-1}}{\text{minimize}} \quad 
    d\!\left(\mathcal{T}(s_s, x_K),\, \mathcal{T}(s_s, s_g)\right)
    \label{eq:objective}
\end{equation}
where $d(\cdot,\cdot)$ measures the remaining discrepancy using task-specific visual, geometric, or scene-state signals. Under this formulation, we instantiate four controlled and verifiable editing families: object placement, which edits object pose through translation and rotation; shape-key editing, which adjusts blend-shape coefficients to match target geometry while preserving object identity; lighting adjustment, which modifies illumination attributes, primarily color and energy, to match target lighting effects; and material modification, which updates shader parameters such as base color, roughness, metallic, and alpha for appearance transfer.

\subsection{Scalable 3D Graphics Editing Data Generation}
\label{sec:data_generation}
\begin{figure}[t]
    \centering
    \includegraphics[width=1.0\textwidth]{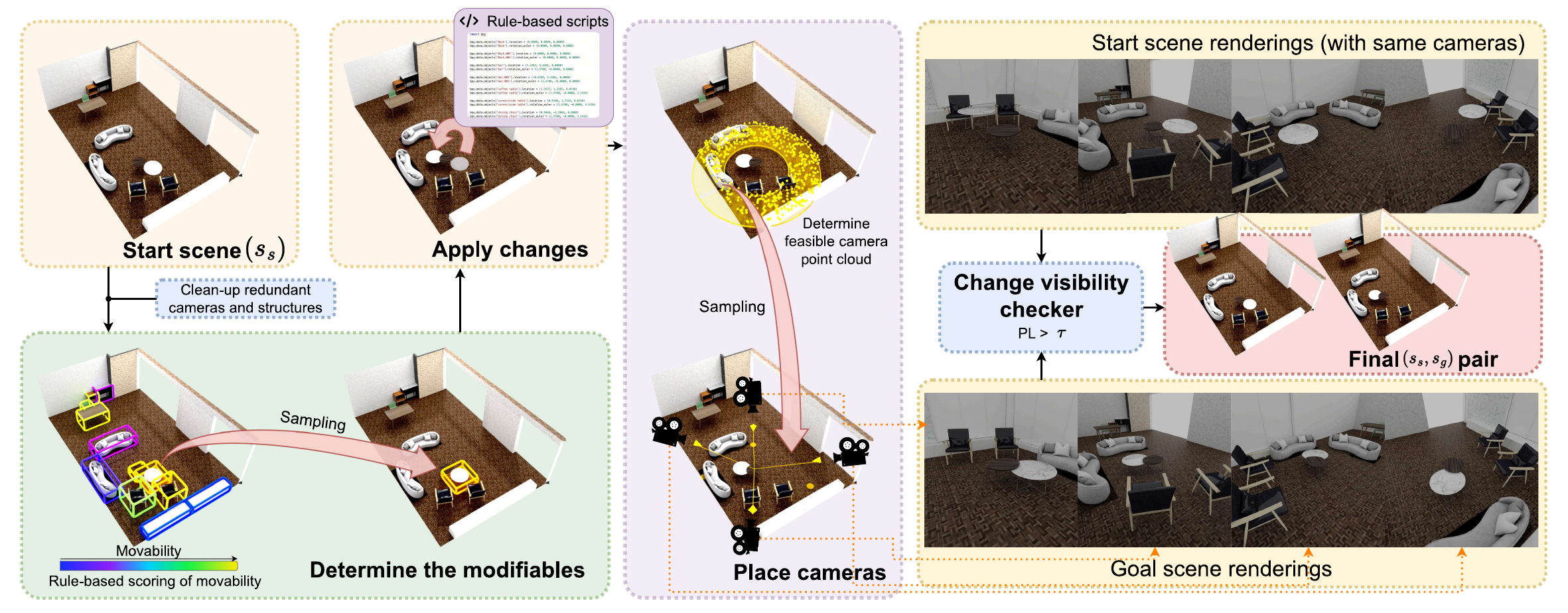}
    \caption{\textbf{Perturbation-based data generation pipeline.}
    We illustrate the pipeline using object placement as an example. Starting from a base Blender scene, the pipeline removes redundant scene elements, selects editable objects or attributes, applies rule-based perturbations, and places aligned cameras so that the start--goal change remains observable. The resulting start and goal renderings are passed through a visibility check before the final $(s_s, s_g)$ task pair is exported.}
    \label{fig:data_pipeline}
\end{figure}

A key bottleneck for training executable 3D editing agents is the scarcity of structured editing supervision. Although 3D assets and scenes are increasingly available, they are rarely paired with start and goal states, language instructions, editable attributes, aligned observations, and verification signals. We therefore build a perturbation-based data pipeline that turns Blender assets into start-to-goal editing tasks for supervised training, RL rollout environments, and reward computation.

We draw base assets from both generated and curated sources. For object placement, we sample 1000 3D-FRONT scenes~\cite{fu20213d,fu20213dfuture}, partitioning multi-room scenes when necessary. For lighting adjustment, we use 226 Infinigen scenes~\cite{raistrick2024infinigen}, which naturally contain lamp meshes and light objects. For shape-key editing, we curate 54 assets with 409 modifiable shape keys, and use Qwen3-14B~\cite{bai2025qwen3} to rename keys during base-scene construction, yielding over 4000 shape-key variants. For material modification, we curate 28 Infinigen material definitions covering over 100 assets and build a static material collection by replacing stochastic parameters with preset values. We also include Blender's Principled BSDF materials for controlled material edits.

For each base asset, the pipeline samples a start state $s_s$ and a goal state $s_g$ by perturbing task-specific editable attributes. Object-placement tasks perturb selected object translations and rotations under scene-boundary and collision constraints. Shape-key tasks perturb selected blend-shape coefficients while preserving object identity. Lighting tasks adjust illumination color or intensity, while material tasks alter Blender shader parameters such as base color, roughness, metallic, and alpha, as well as attributes exposed by Infinigen material definitions. The resulting state difference defines the editing instruction and provides supervision signals for downstream SFT and RL training.

The generation procedure follows a common structure while preserving task-specific constraints. For placement, the pipeline identifies movable objects by filtering room structures and abnormal objects, then ranks candidates using geometric and semantic mobility cues before sampling valid pose perturbations. For lighting, the pipeline selects editable light attributes and places cameras to make the adjusted illumination observable. Shape-key and material scenes use curated cameras and predefined editable attributes, which reduces ambiguity in fine-grained geometry and appearance edits. This design gives the four task families a shared start-to-goal interface while preserving the constraints needed for reliable data generation.

Finally, generated tasks are filtered for observability and reproducibility. Placement and lighting tasks use aligned start-goal camera views so that the relevant perturbations remain visible. Samples with invalid collisions or insufficient visual change are rejected or resampled. For material tasks, which may introduce stochastic rendering noise, we additionally repeat rendering checks and keep only visually stable samples. Each accepted task is exported with start/goal scenes, aligned multi-view renderings, task instructions, and structured scene descriptors. Detailed perturbation ranges, camera-placement rules, and split configurations are provided in the Appendix.

\subsection{Environment: Blender Sandbox}
\label{sec:sandbox}

A central requirement for training 3D editing agents is an execution environment in which model-generated actions induce reproducible scene transitions and yield measurable feedback. We therefore design the \emph{Blender Sandbox}, an instance-isolated environment that exposes Blender Python (\texttt{bpy}) execution, multi-view rendering, and structured scene-state export through a unified interaction interface. In contrast to setups that treat Blender as a one-off subprocess~\cite{Gu_2025_CVPR}, the sandbox is used consistently across trajectory generation, reinforcement-learning rollouts, and evaluation, so that the agent is optimized and tested under the same transition and observation semantics.

The sandbox is built around three design principles.

\noindent\textbf{Instance-Level Isolation.} Each episode runs in a dedicated Blender process spawned upon task initialization. Scene state, global variables, and renderer parameters are never shared across episodes, ensuring that each reinforcement-learning rollout begins from a clean, reproducible initial condition.

\noindent\textbf{Determinism.} The sandbox controls scene initialization and rendering settings to make scene transitions and reward computation reproducible. Rendering consistency is checked within task-specific tolerances, as described in the Appendix.

\noindent\textbf{State Verifiability.} It provides programmatic access to the full scene graph: object poses, shape-key values, material parameters, and lighting attributes, through structured scene state export. This enables quantitative, metric-based task score computation without relying entirely on visual outputs.

\noindent Architecture and implementation details are provided in Appendix~\ref{sec:supp_sandbox}.

\subsection{Evaluation Protocol}
\label{sec:benchmarks}

\subsubsection{Tasks and Datasets.}
We evaluate on the four editing categories defined 
in Sec.~\ref{sec:task_definition} (Placement, Shape, 
Lighting, and Material). 
Following BlenderGym~\cite{Gu_2025_CVPR}, the context input for each 
evaluation episode consists of: \textbf{(i)}~textual information 
describing the start scene $s_s$; \textbf{(ii)}~a set of 
multi-view renderings of $s_s$ with camera identifiers; and 
\textbf{(iii)}~corresponding renderings from the goal scene 
$s_g$. During editing, the model interacts with the Blender Sandbox
environment through two tools: \texttt{blender\_code\_executor} and 
\texttt{blender\_preview\_renderer} (Sec.~\ref{sec:sandbox}), in a ReAct-style Tool Agent Loop~\cite{yao2022react, schick2023toolformer}. Each task family contains 200 test tasks. Episodes interrupted by API timeouts or server errors are excluded from both numerator and denominator, so a model may have fewer than 200 evaluated episodes per family; model-generated code errors instead allow further iteration within the episode. Dataset splits and prompts are detailed in Appendix~\ref{sec:supp_data_prompts}.

\subsubsection{Evaluation Metrics.}
\label{sec:eval_metrics}

We define a unified distance-based framework for measuring editing quality. This metric is used for both evaluation benchmarking and reward computation (Sec.~\ref{sec:rl}).

\noindent\textbf{Distance Functions.}
We employ task-specific distance functions to quantify discrepancy between scene states.
Following prior work~\cite{Gu_2025_CVPR,huang2024blenderalchemyediting3dgraphics}, we adopt Photometric Loss (PL) and Negative-CLIP Distance (one minus the CLIP similarity, denoted as N-CLIP) as visual distances between renderings. 
For the \emph{Placement} task, we propose the Part-aware Chamfer Distance (P-CD), which computes the Chamfer Distance between points sharing the same semantic label and averages across all $|\mathcal{P}|$ parts:
\begin{equation}
    \text{P-CD}(s_i, s_j) = \frac{1}{|\mathcal{P}|}\sum_{k=1}^{|\mathcal{P}|}\text{CD}(s_i^k, s_j^k),
\end{equation}
where $s^k$ denotes the point cloud of the $k$-th semantic part.

For scene-state comparison, we compute task-specific distances from structured scene descriptors exported by the sandbox (Sec.~\ref{sec:sandbox}): placement decomposes into object-identification distance (one minus Jaccard similarity over moved vs.\ ground-truth objects) and pose distance (weighted Euclidean-geodesic distance); shape-key tasks use the ShapeKey-L1 distance, defined as the L1 norm between shape-key value vectors; material tasks use vectorized parameter L1 distances; 
and lighting employs color and energy difference.
To simplify presentation, we denote distance from state $s_i$ to $s_j$ as $d_{i, j}$.

\noindent\textbf{Progress score.}
To compare edits across scenes and task families with different distance scales, we normalize each episode by its initial start-goal discrepancy:
\begin{equation}
    \mathrm{PS}_{D}(s_m;\,s_s,\,s_g)
    =
    \mathrm{clip}
    \left(
    \frac{D(s_s,s_g)-D(s_m,s_g)}
    {\max(D(s_s,s_g),\epsilon)},
    0,1
    \right),
    \label{eq:progress_score}
\end{equation}
where $\epsilon$ is a small constant for numerical stability.
A score of $0$ indicates no measured improvement over the start scene, while a score of $1$ indicates alignment with the goal under distance $D$.
Edits that move away from the goal are clipped to zero progress.

\noindent\textbf{Success Rate.} Given a threshold $\tau$, we define SR@$\tau$ as:
\begin{equation}
    \text{SR@}\tau = P\!\left( \frac{d_{m,g}}{d_{s,g}} \leq \tau \right),
\end{equation}
where $d$ is P-CD for placement and ShapeKey-L1 for shape editing.
We additionally report the median normalized residual for both tasks.

\subsection{Training}
\label{sec:training}

\subsubsection{Stage 1: Supervised Fine-Tuning.}

We initialize from the open-weight Qwen3-VL-8B-Thinking model~\cite{bai2025qwen3} and perform supervised fine-tuning (SFT) with the standard autoregressive next-token prediction objective on assistant tokens.
This stage provides a cold start for 3D editing by teaching executable tool-use format, Blender Python editing patterns, visual feedback interpretation, and task-specific scene reasoning, while preserving general reasoning ability.
Our SFT mixture contains two complementary sources.

\noindent\textbf{Trajectory-SFT.}
A strong proprietary MLLM serves as the teacher and interacts with the Blender Sandbox across all task categories in a ReAct-style tool-agent loop~\cite{yao2022react, schick2023toolformer}.
As shown in the \emph{Trajectory-SFT} block of Fig.~\ref{fig:main_graph}, the collected data are multi-turn trajectories comprising reasoning traces, tool-call requests, and multimodal tool responses.
We filter these trajectories using task-specific metrics aligned with the reward functions in Sec.~\ref{sec:state_export}.

\noindent\textbf{General-SFT.}
To mitigate catastrophic forgetting~\cite{dong2024abilities}, we mix general-purpose data with task trajectories at a $36{:}1$ ratio.
The general-purpose data include:
\textbf{(i)} VQA samples that ask whether an auxiliary state $s_c$ is closer to the goal $s_g$ than the start state $s_s$, labeled by $\mathbb{I}(D(s_c,s_g)<D(s_s,s_g))$;
\textbf{(ii)} function-calling examples that preserve structured tool-use formatting; and
\textbf{(iii)} coding problems that preserve code-generation ability.
Together, these data maintain multimodal instruction following, tool-use syntax, and programming priors needed for interactive editing.

\subsubsection{Stage 2: AgentLoop-RL.}
\label{sec:rl}

After SFT provides an executable cold-start policy, AgentLoop-RL further optimizes the agent through closed-loop interaction with the Blender Sandbox.
We use GSPO~\cite{zheng2025group} for reinforcement learning over long, tool-augmented trajectories.

\noindent\textbf{RL formulation.}
We formulate tool-agent training as an autoregressive Markov Decision Process (MDP) $\langle \mathcal{S}, \mathcal{A}, \mathcal{P}, \mathcal{R}, \gamma \rangle$.
At turn $t$, the state $s_t$ contains the initial task context and prior model outputs and tool responses.
The action $a_t \sim \pi_\theta(\cdot|s_t)$ is a sequence-level macro-action that may include reasoning text and executable tool calls.
The Blender Sandbox mediates the transition: code execution updates the scene, preview rendering returns visual observations, and the dialogue state becomes $s_{t+1}=s_t\oplus a_t\oplus z_t$, where $z_t$ is the tool response.
This environment-mediated transition makes the final scene state directly measurable for reward computation.

\noindent\textbf{Progress-normalized verifiable rewards.}
\label{sec:state_export}
At the end of each episode, the sandbox exports the final modified scene $s_m$, multi-view renderings, structured scene descriptors, and point-cloud samples when required.
Using the progress score in Eq.~\ref{eq:progress_score}, we compute three complementary progress components: visual progress $r_{\mathrm{vis}}$, scene-state progress $r_{\mathrm{state}}$, and geometric progress $r_{\mathrm{geo}}$.
The visual component is averaged over multi-view PL or N-CLIP distances; the scene-state component is computed from task-specific structured descriptors and penalizes unintended attribute changes; and the geometric component is computed from P-CD for Placement.
The task-progress reward $r_{\mathrm{prog}}(s_m)$ is a task-specific weighted average of these components; the per-family configuration is given in Sec.~\ref{sec:exp_setup}.

\noindent\textbf{Trajectory-level actionability signal.}
Tool-augmented RL faces a sparse exploration problem: task-progress rewards become informative only after the policy enters the executable action space and induces an environment transition.
We therefore incorporate a bounded auxiliary signal $\mathcal{A}(\tau)\in[0,1]$ for trajectory $\tau$, indicating whether the episode contains valid executable tool use and produces a measurable scene-state transition.
This signal eases exploration by rewarding valid editing actions; task success is scored only by $r_{\mathrm{prog}}$.

\noindent\textbf{Final reward.}
The scalar reward is:
\begin{equation}
    r(\tau)
    =
    (1-\lambda_{\mathrm{act}}) r_{\mathrm{prog}}(s_m)
    +
    \lambda_{\mathrm{act}} \mathcal{A}(\tau),
    \quad
    0 \leq \lambda_{\mathrm{act}} \ll 1.
    \label{eq:reward}
\end{equation}
Because the auxiliary term is at most $\lambda_{\mathrm{act}}$, high reward requires task progress.

\section{Experiments}
\label{sec:experiments}

\begin{table*}[t!]
\centering
\caption{
    Comparison of models on the test split of \textbf{BlenderFORGE} (metrics in Sec.~\ref{sec:eval_metrics}).
    For \textit{Placement} and \textit{Shape}, SR@$\tau$ (\%) denotes success rate at threshold $\tau$ (higher is better);
    Med.\ denotes the median discrepancy (lower is better).
    PL ($\times10^{-3}$) and N-CLIP ($\times10^{-3}$) measure perceptual quality (lower is better). 
    The w/o change row reports the discrepancy between the start scene $s_s$ and the goal scene $s_g$. 
    Within the \textbf{Proprietary Models} block and the combined \textbf{Open-weight Models}/\textbf{Fine-tuned} blocks, the best performance is marked with \textbf{bold}, and the second-best distinct score is \underline{underlined}. 
    The last two rows show gains of each BlenderFORGE model over its corresponding base: increases in SR (percentage points) and decreases in distance metrics.
    The 8B base Placement results use the $N=200$ evaluation in Tab.~\ref{tab:ablation}.
}
\label{tab:main_results}
\resizebox{\textwidth}{!}{%
\begin{tabular}{l | cccccc | cccccc | cc | cc}
\toprule
& \multicolumn{6}{c|}{\textbf{Placement}}
& \multicolumn{6}{c|}{\textbf{Shape}}
& \multicolumn{2}{c|}{\textbf{Lighting}}
& \multicolumn{2}{c}{\textbf{Material}} \\
\cmidrule(lr){2-7}\cmidrule(lr){8-13}\cmidrule(lr){14-15}\cmidrule(lr){16-17}
& \multicolumn{2}{c}{Perceptual}
& \multicolumn{4}{c|}{Normalized P-CD}
& \multicolumn{2}{c}{Perceptual}
& \multicolumn{4}{c|}{Normalized ShapeKey L1}
& \multicolumn{2}{c|}{Perceptual}
& \multicolumn{2}{c}{Perceptual} \\
\cmidrule(lr){2-3}\cmidrule(lr){4-7}\cmidrule(lr){8-9}\cmidrule(lr){10-13}\cmidrule(lr){14-15}\cmidrule(lr){16-17}
\textbf{Model}
& PL$\downarrow$ & N-CLIP$\downarrow$
& SR@.1$\uparrow$ & SR@.3$\uparrow$ & SR@.5$\uparrow$ & Med.$\downarrow$
& PL$\downarrow$ & N-CLIP$\downarrow$
& SR@.1$\uparrow$ & SR@.3$\uparrow$ & SR@.5$\uparrow$ & Med.$\downarrow$
& PL$\downarrow$ & N-CLIP$\downarrow$
& PL$\downarrow$ & N-CLIP$\downarrow$ \\
\midrule
w/o change
    & 6.4 & 23.5
    & 0.00 & 0.00 & 0.00 & 1.00
    & 5.2 & 32.0
    & 0.00 & 0.00 & 0.00 & 1.00
    & 43.8 & 38.2
    & 16.0 & 76.0 \\
\midrule
\multicolumn{17}{l}{\textit{Proprietary Models}} \\
\midrule
GPT-5.2
    & \textbf{4.4} & \textbf{15.9}
    & \textbf{27.06} & \textbf{50.43} & \textbf{62.76} & \textbf{0.39}
    & \textbf{3.2} & \textbf{11.2}
    & \textbf{17.28} & \textbf{30.89} & \textbf{39.27} & \textbf{0.71}
    & \textbf{6.6} & \textbf{16.3}
    & 7.1 & \underline{34.1} \\
Claude-Sonnet-4.6
    & 5.9 & \underline{21.8}
    & 10.00 & 25.00 & 32.50 & 0.91
    & \underline{3.6} & \underline{19.4}
    & \underline{13.50} & 26.00 & 35.00 & 0.87
    & \underline{6.9} & \underline{18.3}
    & \textbf{4.6} & 35.2 \\
Gemini-3-Pro
    & 6.4 & 23.6
    & 9.53 & \underline{25.02} & \underline{36.69} & \underline{0.83}
    & 4.4 & 22.6
    & 10.50 & 23.00 & 27.50 & 1.18
    & 15.3 & 23.8
    & 29.2 & 46.0 \\
Gemini-3-Flash
    & \underline{5.8} & 22.1
    & 6.00 & 21.50 & 35.50 & 0.96
    & \underline{3.6} & 19.5
    & 13.00 & \underline{29.50} & \underline{36.50} & \underline{0.85}
    & 12.7 & 25.8
    & 24.4 & 49.4 \\
Gemini-2.5-Pro
    & 6.0 & 23.0
    & \underline{11.50} & 20.00 & 31.00 & 1.00
    & 5.0 & 26.3
    & 7.50 & 18.00 & 22.50 & 1.18
    & 20.9 & 29.2
    & 7.5 & 40.0 \\
Gemini-2.5-Flash
    & 6.3 & 25.6
    & 4.00 & 11.00 & 16.50 & 1.30
    & 5.7 & 27.9
    & 8.50 & 16.00 & 20.50 & 1.09
    & 26.5 & 29.0
    & \underline{6.5} & \textbf{33.4} \\
Claude-Sonnet-3.5
    & 6.8 & 23.1
    & 3.50 & 8.00 & 17.00 & 0.96
    & 5.0  & 28.2
    & 3.50 & 9.50    & 16.50 & 1.03
    & 32.2 & 31.4
    & 7.5 & 53.6 \\
Qwen3.5-Plus
    & 6.0 & 23.7
    & 5.78 & 16.40 & 25.48 & 1.47
    & 4.3 & 23.7
    & 10.50 & 24.50 & 30.50 & 0.99
    & 20.6 & 31.1
    & 9.9 & 44.3 \\
Qwen3-VL-Plus
    & 8.2 & 32.2
    & 1.14 & 4.00  & 5.71  & 2.53
    & 6.2 & 28.5
    & 7.50 & 15.00 & 21.00 & 1.53
    & 34.2 & 35.0
    & 16.8 & 64.6 \\
\midrule
\multicolumn{17}{l}{\textit{Open-weight Models}} \\
\midrule
\begin{tabular}{@{}l@{}}GLM-4.6V (106B)\end{tabular}
    & 9.6 & 29.9
    & 2.00 & 7.00 & 9.50 & 2.63
    & 5.5 & \underline{27.0}
    & 7.00 & \underline{15.50} & \underline{24.00} & 1.12
    & 30.4 & 31.6
    & \underline{8.1} & \underline{44.0} \\
\begin{tabular}{@{}l@{}}Mistral-Small-3.2 (24B)\end{tabular}
    & 7.3 & 26.2
    & 1.01 & 2.01 & 2.01 & 1.00
    & 6.5 & 35.9
    & 2.00 & 5.00 & 7.00 & \textbf{1.00}
    & 45.2 & 37.7
    & 11.7 & 62.6 \\
\begin{tabular}{@{}l@{}}InternVL3 (8B)\end{tabular}
    & 9.4 & 31.8
    & 1.50 & 3.00 & 4.50 & 1.14
    & 5.7 & 34.2
    & 0.00 & 0.00 & 0.50 & \textbf{1.00}
    & 44.2 & 38.3
    & 43.2 & 88.4 \\
\begin{tabular}{@{}l@{}}Llama4-Maverick (400B)\end{tabular}
    & 11.0 & 32.6
    & 0.50 & 1.00 & 1.00 & 1.18
    & 5.3 & 30.2
    & 3.54 & 6.57 & 9.60 & \textbf{1.00}
    & 48.4 & 43.3
    & 23.2 & 82.9 \\
\begin{tabular}{@{}l@{}}Qwen3-VL-8B-Thinking\end{tabular}
    & 18.8 & 57.1
    & 0.00 & 1.50 & 1.50 & 21.31
    & 6.1 & 30.0
    & 4.50 & 9.00 & 12.00 & 1.31
    & 43.0 & 37.7
    & 23.9 & 70.5 \\
\begin{tabular}{@{}l@{}}Qwen3.5-9B\end{tabular}
    & 7.8 & 30.0
    & 0.50 & 1.50 & 3.00 & 1.00
    & 8.4 & 39.8
    & 2.50 & 4.00 & 6.00 & \textbf{1.00}
    & 36.1 & 39.9
    & 16.9 & 54.8 \\
\midrule
\multicolumn{17}{l}{\textit{Fine-tuned on BlenderFORGE Datasets}} \\
\midrule
\begin{tabular}{@{}l@{}}Qwen3-VL-8B\\(BlenderFORGE)\end{tabular}
    & \underline{6.1} & \underline{22.5}
    & \underline{8.50} & \underline{24.00} & \underline{36.00} & \underline{0.78}
    & \textbf{4.7} & \textbf{23.0}
    & \underline{8.50} & \underline{15.50} & 20.00 & \underline{1.07}
    & \underline{17.1} & \underline{25.4}
    & 12.7 & 53.6 \\
\begin{tabular}{@{}l@{}}Qwen3.5-9B\\(BlenderFORGE)\end{tabular}
    & \textbf{5.7} & \textbf{20.5}
    & \textbf{15.50} & \textbf{29.50} & \textbf{40.00} & \textbf{0.73}
    & \underline{5.1} & 28.3
    & \textbf{10.50} & \textbf{21.00} & \textbf{27.50} & \textbf{1.00}
    & \textbf{7.3} & \textbf{19.8}
    & \textbf{7.8} & \textbf{34.0} \\
\midrule
\begin{tabular}{@{}l@{}}\color{green!70!black}8B Improvement \\ \color{green!70!black}over Base\end{tabular}
    & \color{green!70!black}12.7 & \color{green!70!black}34.6
    & \color{green!70!black}8.50 & \color{green!70!black}22.50 & \color{green!70!black}34.50 & \color{green!70!black}20.53
    & \color{green!70!black}1.4   &  \color{green!70!black}7.0
    & \color{green!70!black}4.00   & \color{green!70!black}6.50 & \color{green!70!black}8.00    & \color{green!70!black}0.24
    & \color{green!70!black}25.9   &  \color{green!70!black}12.3
    & \color{green!70!black}11.2   &  \color{green!70!black}16.9  \\
\begin{tabular}{@{}l@{}}\color{green!70!black}9B Improvement \\ \color{green!70!black}over Base\end{tabular}
    & \color{green!70!black}2.1 & \color{green!70!black}9.5
    & \color{green!70!black}15.00 & \color{green!70!black}28.00 & \color{green!70!black}37.00 & \color{green!70!black}0.27
    & \color{green!70!black}3.3 & \color{green!70!black}11.5
    & \color{green!70!black}8.00 & \color{green!70!black}17.00 & \color{green!70!black}21.50 & 0.00
    & \color{green!70!black}28.8 & \color{green!70!black}20.1
    & \color{green!70!black}9.1 & \color{green!70!black}20.8 \\
\bottomrule
\end{tabular}%
}
\end{table*}

We evaluate BlenderFORGE-trained agents against their open-weight base models under the protocol in Sec.~\ref{sec:benchmarks}, with proprietary MLLMs serving as zero-shot reference points for benchmark difficulty. Additional evaluations examine a second backbone and transfer.

\subsection{Model and Training Setup}
\label{sec:exp_setup}

We use Qwen3-VL-8B-Thinking~\cite{bai2025qwen3} as the open-weight baseline model and train it with the two-stage pipeline described in Sec.~\ref{sec:training}. During SFT, we fine-tune only the language model while freezing the vision encoder and multimodal projector, using Swift~\cite{zhao2025swift}. We jointly train a single agent on a shuffled mixture of editing trajectories from all four task families, generated by the teacher GPT-5.2~\cite{openai2025gpt52}, and then run 240 steps of General-SFT. Both SFT stages use a learning rate of $1\times10^{-5}$ and batch size 64.

For AgentLoop-RL, we initialize from the SFT checkpoint and train with VeRL/HybridFlow~\cite{sheng2025hybridflow} using GSPO~\cite{zheng2025group}. Rewards are computed by the Blender Sandbox from the progress-normalized verification signals in Sec.~\ref{sec:state_export}. For each task family, we use the verification signal most directly tied to the edited attribute: geometric progress for placement, scene-state and visual progress for shape-key editing, and visual progress for lighting and material editing. The same task-specific reward configuration is shared by the ablation variants. We use learning rate $1\times10^{-6}$, batch size 2, maximum context length 32,768, and a maximum of $T=8$ interaction turns during RL training. The KL coefficient is set to 0 to allow exploration during closed-loop editing. Training uses eight NVIDIA RTX 5090 GPUs for model optimization; up to eight additional GPUs are assigned to sandbox rendering. We also train Qwen3.5-9B with the same pipeline, likewise as a single jointly trained agent.

\subsection{Experiment Results and Analysis}
\label{sec:exp_main}

In Tab.~\ref{tab:main_results}, we benchmark 15 zero-shot MLLMs and our fine-tuned agents, including \textbf{(i)} 9 representative proprietary models with strong multimodal understanding and agentic capabilities, \textbf{(ii)} 6 open-weight MLLMs, and \textbf{(iii)} BlenderFORGE-trained Qwen3-VL-8B and Qwen3.5-9B agents~\cite{bai2025qwen3,qwen35}. The fine-tuned models are trained on the BlenderFORGE train split, with 3,000 samples for object placement and 2,000 samples for each remaining task family. To keep the comparison aligned with each model's intended inference mode, we enable thinking for the Qwen series~\cite{qwen35,bai2025qwen3}, and use the strongest available reasoning setting for the Gemini series~\cite{comanici2025gemini,google2025gemini3promodelcard} and GPT-5.2~\cite{singh2025openaigpt5card}. We also report the ``w/o change'' row, which measures the start-to-goal discrepancy before any edit and indicates whether an agent reduces the task discrepancy compared with leaving the scene unchanged.

\noindent\textbf{Behavior of zero-shot MLLM agents.}
Executable 3D graphics editing exposes failure modes that are shared across current zero-shot MLLM agents rather than tied to a single model family. From Tab.~\ref{tab:main_results}, we observe that many models improve some perceptual metrics but do not consistently reduce the task discrepancy across all editing families, and some underperform the ``w/o change'' reference on certain metrics. We identify two recurring patterns. \textbf{(i)}~\emph{Inaccurate multi-view 3D grounding.} In placement tasks, agents must map differences between aligned start/goal renderings to object identities, translations, and rotations in the Blender scene. Zero-shot agents often rely on coarse visual estimates or trial-and-error feedback, rather than reliably inferring the underlying 3D transformation across views. This is especially challenging when the required edit must be expressed in Blender's scene coordinate frame rather than in a camera-local view. \textbf{(ii)}~\emph{Interaction-context drift.} As tool outputs and preview renderings accumulate, agents may lose track of the original goal, hallucinate intermediate observations, or terminate early while visible residual differences remain. These patterns suggest that current MLLMs can invoke Blender tools but do not yet reliably complete closed-loop 3D edits in the zero-shot setting. More qualitative examples are presented in the Appendix.

\begin{figure}[t!]
    \centering
    \includegraphics[width=1.0\textwidth]{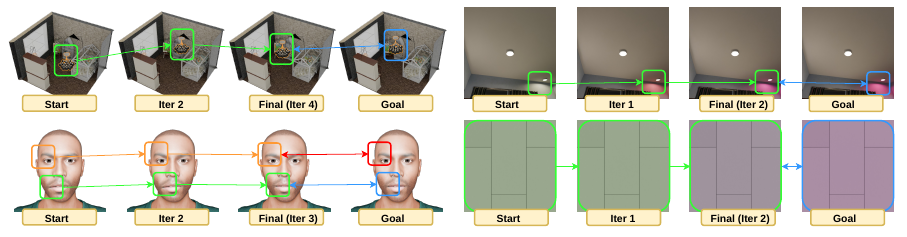}
    \caption{\textbf{Qualitative results of Qwen3-VL-8B-BlenderFORGE}. {\color{green!70!black}Green} and {\color{blue}blue} traces indicate positive progress, while {\color{orange}orange} and {\color{red}red} traces indicate negative progress. Overall, the model improves the scene through fine-grained edits, although some detailed errors remain.}
    \label{fig:Qualitative}
\end{figure}

\noindent\textbf{Effect of BlenderFORGE training.}
We next examine whether BlenderFORGE training improves open-weight MLLM agents for executable 3D graphics editing. In Tab.~\ref{tab:main_results}, Qwen3-VL-8B-BlenderFORGE improves over its base model (Qwen3-VL-8B-Thinking) on all four task families and all reported metrics. The improvements are especially pronounced for placement and lighting, where the agent must connect visual differences with executable scene updates. Trained with the same pipeline, Qwen3.5-9B (Sec.~\ref{sec:transfer}) ranks behind only GPT-5.2 among zero-shot models on all six Placement metrics, and behind only GPT-5.2 and Claude-Sonnet-4.6 on both Lighting metrics. Although the trained agents still trail the strongest proprietary models such as GPT-5.2 and Claude-Sonnet-4.6 on several metrics, they substantially improve over their open-weight base models and become competitive with several zero-shot and open-weight baselines on selected metrics, suggesting that the proposed data generation, sandbox interaction, and two-stage training pipeline can amortize editing experience into model parameters.

\subsection{Ablation Studies}
\label{sec:ablation}
On the Placement task, Tab.~\ref{tab:ablation} compares Trajectory-SFT, General-SFT, and AgentLoop-RL on a fixed $N=200$ split, using the same Base results as Tab.~\ref{tab:main_results}. We draw three observations.

\noindent\textbf{Observation 1. }All training configurations improve Placement edit quality over Base (SR@.5: 8.50--36.00\% versus 1.50\%). These results support the value of teacher trajectories and verifiable-reward training in this task.

\noindent\textbf{Observation 2. }General-SFT reduces average interaction turns from 8.01 to 5.76 without RL and from 10.63 to 6.49 with RL, while increasing Placement SR@.5 in both comparisons.

\noindent\textbf{Observation 3. }The full pipeline achieves the best Placement edit-quality metrics, with SR@.5 of 36.00\% versus 25.50\% for the next-best variant. RL is not uniformly beneficial: Traj+RL reaches 21.00\%, below Traj SFT at 24.00\%. Traj+Gen uses the fewest turns; the full pipeline offers higher edit quality at additional interaction cost.

\begin{table}[t!]
\centering
\caption{
    Ablation study on the \textit{Placement} task with $N=200$ for every row.
    SR@$\tau$ (\%) denotes success rate at threshold $\tau$ (higher is better);
    Med.\ denotes the median P-Chamfer distance (lower is better).
    PL ($\times10^{-3}$) and N-CLIP ($\times10^{-3}$) measure perceptual quality (lower is better).
    Avg.\ turns denotes the average number of interaction turns.
    {\color{green!70!black}Green} deltas indicate improvement over the base model.
}
\label{tab:ablation}

\resizebox{\columnwidth}{!}{%
\begin{tabular}{ccc | cc | cccc | c}
\toprule
\multicolumn{3}{c|}{\textbf{Training Stage}} 
& \multicolumn{2}{c|}{\textbf{Perceptual}} 
& \multicolumn{4}{c|}{\textbf{Normalized P-CD}} 

& \multirow{2}{*}{\makecell{\textbf{Avg.}\\ \textbf{turns}}} \\ 
\cmidrule(lr){1-3}\cmidrule(lr){4-5}\cmidrule(lr){6-9}

Traj. & Gen. & RL 
& PL$\downarrow$ & N-CLIP$\downarrow$ 
& SR@.1$\uparrow$ & SR@.3$\uparrow$ & SR@.5$\uparrow$ & Med.$\downarrow$ 
& \\ 
\midrule

 &  &  & 18.8 & 57.1 & 0.00 & 1.50 & 1.50 & 21.31 & 9.29 \\
 &  & \checkmark & 10.6 {\color{green!70!black}(8.2)} & 33.3 {\color{green!70!black}(23.8)} & 2.00 {\color{green!70!black}(2.00)} & 5.00 {\color{green!70!black}(3.50)} & 8.50 {\color{green!70!black}(7.00)} & 1.64 {\color{green!70!black}(19.67)} & 10.08 \\
\checkmark &  &  & 6.5 {\color{green!70!black}(12.3)} & 24.1 {\color{green!70!black}(33.0)} & 4.50 {\color{green!70!black}(4.50)} & 13.00 {\color{green!70!black}(11.50)} & 24.00 {\color{green!70!black}(22.50)} & 1.00 {\color{green!70!black}(20.31)} & 8.01 \\
\checkmark & \checkmark &  & 6.9 {\color{green!70!black}(11.9)} & 24.0 {\color{green!70!black}(33.1)} & 6.50 {\color{green!70!black}(6.50)} & 17.50 {\color{green!70!black}(16.00)} & 25.50 {\color{green!70!black}(24.00)} & 1.01 {\color{green!70!black}(20.30)} & \textbf{5.76} \\
\checkmark &  & \checkmark & 6.5 {\color{green!70!black}(12.3)} & 24.9 {\color{green!70!black}(32.2)} & 3.00 {\color{green!70!black}(3.00)} & 14.00 {\color{green!70!black}(12.50)} & 21.00 {\color{green!70!black}(19.50)} & 1.03 {\color{green!70!black}(20.28)} & 10.63 \\
\checkmark & \checkmark & \checkmark & \textbf{6.1} {\color{green!70!black}(12.7)} & \textbf{22.5} {\color{green!70!black}(34.6)} & \textbf{8.50} {\color{green!70!black}(8.50)} & \textbf{24.00} {\color{green!70!black}(22.50)} & \textbf{36.00} {\color{green!70!black}(34.50)} & \textbf{0.78} {\color{green!70!black}(20.53)} & 6.49 \\
\bottomrule
\end{tabular}%
}
\end{table}

\subsection{Second Backbone and Transfer}
\label{sec:transfer}
\begin{table}[htbp]
\centering
\small
\setlength{\tabcolsep}{4pt}
\caption{Qwen3.5-9B transfer to BlenderGym. PL and N-CLIP use the $\times10^{-3}$ convention; SR values are percentages. Dashes: SR and Med.\ are defined only for Placement and Shape-Key.}
\label{tab:blendergym_transfer}
\begin{tabular}{llrrrr}
\toprule
Task & Model & PL $\downarrow$ & N-CLIP $\downarrow$ & SR@.5 $\uparrow$ & Med. $\downarrow$ \\
\midrule
\multirow{2}{*}{Placement} & Base & 8.8 & 34.4 & 20.00 & 1.00 \\
 & BlenderFORGE & 8.4 & 30.3 & 36.00 & 0.91 \\
\midrule
\multirow{2}{*}{Shape-Key} & Base & 12.2 & 31.9 & 10.26 & 1.00 \\
 & BlenderFORGE & 8.9 & 26.6 & 17.95 & 0.88 \\
\midrule
\multirow{2}{*}{Lighting} & Base & 52.6 & 49.3 & -- & -- \\
 & BlenderFORGE & 19.0 & 25.3 & -- & -- \\
\bottomrule
\end{tabular}
\end{table}

\noindent\textbf{Second backbone.}
BlenderFORGE training also generalizes to a second backbone: in the fine-tuned block of Tab.~\ref{tab:main_results}, Qwen3.5-9B (BlenderFORGE) improves over its base model on every metric except Shape-Key Med.\ (tied at 1.00), raising Placement SR@.5 from 3.00\% to 40.00\% and Shape-Key SR@.5 from 6.00\% to 27.50\%, and reducing Lighting PL from 36.1 to 7.3.

\noindent\textbf{Transfer to BlenderGym.}
On the independent BlenderGym benchmark~\cite{Gu_2025_CVPR}, the trained Qwen3.5-9B also improves over its base model on all reported metrics (Tab.~\ref{tab:blendergym_transfer}): Placement SR@.5 rises from 20.00\% to 36.00\%, Shape-Key SR@.5 from 10.26\% to 17.95\%, and Lighting PL falls from 52.6 to 19.0. Material is omitted because BlenderGym's legacy Blender version is incompatible with our material-editing API calls (Appendix~\ref{sec:supp_blendergym}).

Results on held-out camera reframing, thinking versus non-thinking baselines, and VLM-based visual assessment are reported in Appendices~\ref{sec:supp_transfer} and~\ref{sec:supp_vlm}.

\subsection{Qualitative Results}
Figure~\ref{fig:Qualitative} presents representative qualitative results.
The trajectories show that the trained agent can iteratively use visual feedback to reduce scene discrepancies across multiple editing steps.
More conversation examples are provided in the Appendix.

\label{sec:qualitative}

\section{Conclusion}
\label{sec:conclusion}

We present BlenderFORGE, a closed-loop framework for optimizing open-weight MLLM agents to perform executable 3D graphics editing through Blender Python code generation. Our framework integrates three key components: a scalable perturbation-based data generation pipeline covering four controlled editing families, an instance-isolated Blender Sandbox that provides deterministic execution and structured feedback, and a two-stage SFT/RL training pipeline for improving agent behavior through offline trajectories and closed-loop rewards. Experiments show that the BlenderFORGE-trained Qwen3-VL-8B and Qwen3.5-9B agents consistently improve over their open-weight base models across all four tasks, narrowing the gap to zero-shot proprietary MLLM baselines and achieving competitive results against several zero-shot or open-weight baselines on selected metrics.

\noindent\textbf{Limitations.}
BlenderFORGE is trained on four state-verifiable editing families. Held-out camera tasks and BlenderGym provide transfer evidence, but camera valid rates reach at most 35\%. Some Shape-Key base meshes are shared across splits despite deduplication. Broader compositional editing and transfer beyond \texttt{bpy} remain open (Appendix~\ref{sec:supp_limitations}).

\begin{ack}
We thank the anonymous reviewers and area chairs for their constructive feedback. This work was supported by the collaborative research initiative between The Hong Kong University of Science and Technology (HKUST) and SenseTime Research, with computational resources provided by SenseTime Research. The authors declare no competing financial interests beyond their respective primary affiliations.
\end{ack}

\bibliography{references}

\newpage
\appendix

\section{Detailed Architecture of the Blender Sandbox}
\label{sec:supp_sandbox}
As mentioned in Sec.~\ref{sec:sandbox} of the main paper, we present the detailed architecture and interaction protocol of the Blender Sandbox. The emphasis here is on the design rationale behind system-level choices that directly impact RL training stability and experimental reproducibility.

\subsection{Architecture and Design Rationale}
\label{sec:supp_arch}

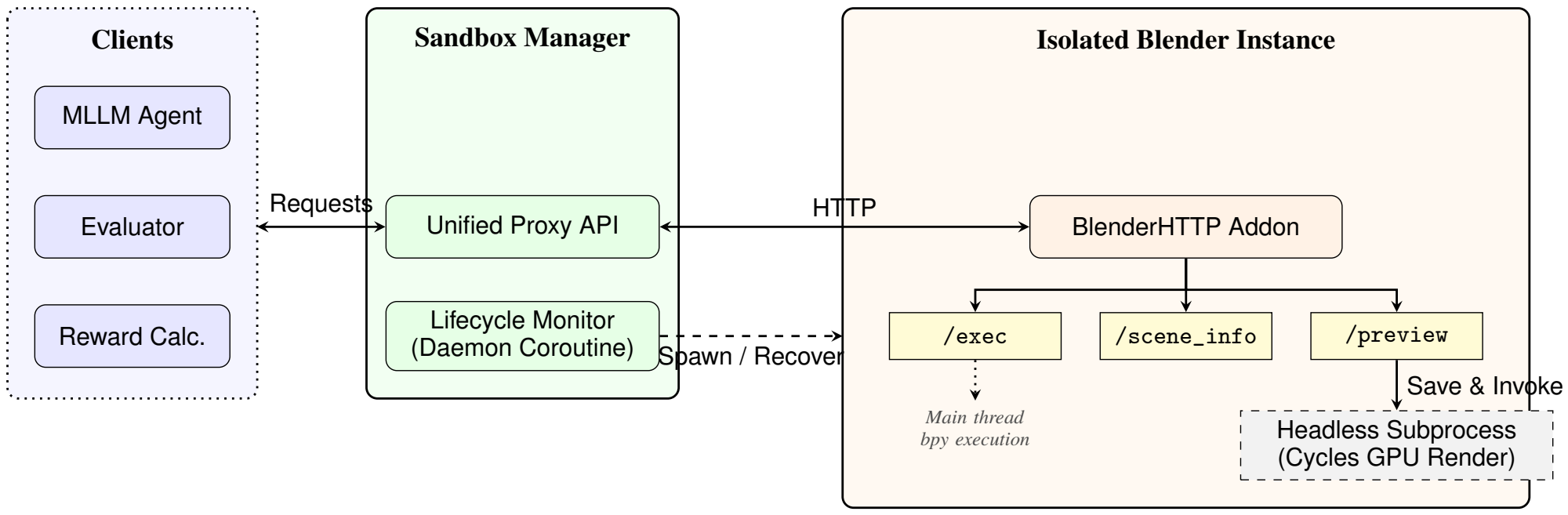
\begin{figure}[h]
    \centering
    \resizebox{0.95\linewidth}{!}{\begin{tikzpicture}[
    font=\sffamily\small,
    >=stealth,
    client/.style={rectangle, draw, rounded corners, fill=blue!10, minimum width=2.5cm, minimum height=0.8cm, align=center},
    manager/.style={rectangle, draw, rounded corners, fill=green!10, minimum width=3.5cm, minimum height=0.8cm, align=center},
    blender/.style={rectangle, draw, rounded corners, fill=orange!10, minimum width=4cm, minimum height=0.8cm, align=center},
    endpoint/.style={rectangle, draw, fill=yellow!20, minimum width=2.2cm, minimum height=0.6cm, align=center},
    subprocess/.style={rectangle, draw, dashed, fill=gray!10, minimum width=4cm, minimum height=0.8cm, align=center}
]

\draw[thick, dotted, fill=blue!4, rounded corners] (-1.6, 1.2) rectangle (1.6, -3.8);
\node[font=\bfseries] at (0, 0.8) {Clients};

\draw[thick, fill=green!5, rounded corners] (3.0, 1.2) rectangle (7.0, -3.8);
\node[font=\bfseries] at (5.0, 0.8) {Sandbox Manager};

\draw[thick, fill=orange!5, rounded corners] (9.1, 1.2) rectangle (17.9, -5.2);
\node[font=\bfseries] at (13.5, 0.8) {Isolated Blender Instance};

\node[client] (agent)  at (0, -0.2) {MLLM Agent};
\node[client] (eval)   at (0, -1.6) {Evaluator};
\node[client] (reward) at (0, -3.0) {Reward Calc.};

\node[manager] (proxy)   at (5.0, -1.6) {Unified Proxy API};
\node[manager] (monitor) at (5.0, -3.0) {Lifecycle Monitor\\(Daemon Coroutine)};

\node[blender]  (http)    at (13.5, -1.6) {BlenderHTTP Addon};
\node[endpoint] (exec)    at (10.8, -3.0) {\texttt{/exec}};
\node[endpoint] (scene)   at (13.5, -3.0) {\texttt{/scene\_info}};
\node[endpoint] (preview) at (16.2, -3.0) {\texttt{/preview}};

\node[subprocess] (headless) at (16.2, -4.4) {Headless Subprocess\\(Cycles GPU Render)};
\node[text width=2.5cm, align=center, font=\scriptsize\itshape, text=black!70] (exec_desc) at (10.8, -4.2) {Main thread\\bpy execution};

\draw[<->, thick] (1.6, -1.6) -- (proxy.west) node[midway, above] {Requests};

\draw[<->, thick] (proxy.east) -- (http.west) node[midway, above] {HTTP};

\draw[->, thick, dashed] (monitor.east) -- (9.1, -3.0) node[midway, below] {Spawn / Recover};

\draw[->, thick] (http.south) -- ++(0,-0.4) -| (exec.north);
\draw[->, thick] (http.south) -- (scene.north);
\draw[->, thick] (http.south) -- ++(0,-0.4) -| (preview.north);

\draw[->, thick] (preview.south) -- (headless.north) node[midway, right] {Save \& Invoke};

\draw[->, dotted, thick] (exec.south) -- (exec_desc.north);

\end{tikzpicture}}
    \caption{Client-server architecture of the Blender Sandbox. The outer \emph{Sandbox Manager} orchestrates isolated Blender instances, each hosting a \emph{BlenderHTTP} addon. Downstream MLLM agents interact exclusively through the proxy API.}
    \label{fig:sandbox_detail}
\end{figure}

The system comprises two layers (Figure~\ref{fig:sandbox_detail}). An outer \emph{Sandbox Manager} service orchestrates the full lifecycle of Blender instances: it dynamically spawns isolated processes on auto-allocated ports, monitors their health, and reclaims idle or crashed instances. Each Blender process hosts an inner \emph{BlenderHTTP} addon---a lightweight HTTP server that exposes three endpoints for code execution, visual preview, and scene-state export. All agent-facing communication is routed through a unified proxy API on the manager, so that downstream clients (MLLM agents, reward calculators, evaluators) are decoupled from internal process management.

Three design decisions are critical to the research contribution and directly support the three principles stated in Sec.~\ref{sec:sandbox} of the main paper: instance-level isolation, determinism, and structured feedback.

\noindent\textbf{(i) Automatic crash recovery for unattended RL training.} Each sandbox instance follows a four-state lifecycle (\textsc{Inactive} $\to$ \textsc{Running} $\to$ \textsc{Killing} $\to$ \textsc{Reinitializing} $\to$ \textsc{Running}). A daemon coroutine continuously monitors the Blender process; upon crash or timeout, it automatically terminates and re-launches the instance without manual intervention. This is essential for the multi-day AgentLoop-RL training described in Sec.~\ref{sec:rl} of the main paper, where Blender segfaults from adversarial agent-generated code would otherwise halt training.

\noindent\textbf{(ii) Deterministic, subprocess-isolated rendering.} The \texttt{/preview} endpoint saves the live scene to a temporary file and invokes a \emph{separate} headless Blender subprocess for GPU-accelerated Cycles rendering with fixed settings (Table~\ref{tab:render_config}). This subprocess isolation serves two purposes: it prevents rendering crashes from corrupting the main session's scene state, and it enforces a canonical render configuration that cannot be altered by agent-generated code, ensuring that the reward signals defined in Sec.~\ref{sec:state_export} of the main paper are computed under identical visual conditions across all episodes. A regex-based safety filter on the code-execution endpoint further blocks operations that could reset the scene or invoke uncontrolled rendering.

\noindent\textbf{(iii) Unified environment across all pipeline stages.} The Blender Sandbox is the \emph{single} back-end for trajectory generation (Sec.~\ref{sec:training}, Stage~1 of the main paper), reinforcement learning (Sec.~\ref{sec:rl}), and evaluation (Sec.~\ref{sec:benchmarks}). The agent always encounters the same API surface, tool schema, rendering pipeline, and error semantics. This eliminates the common pitfall of train--eval distribution mismatch and reduces engineering overhead, as one deployment supports all three workflows without configuration changes.

\begin{table}[h]
    \centering
    \caption{Fixed Cycles render settings enforced for all renderings. These parameters cannot be overridden by the agent, ensuring deterministic visual output across episodes.}
    \label{tab:render_config}
    \begin{tabular}{lc}
        \toprule
        Parameter & Value \\
        \midrule
        Render engine & Cycles (GPU) \\
        Samples / Max bounces & 128 / 8 \\
        Adaptive sampling (threshold) & Enabled (0.2) \\
        Caustics (reflective / refractive) & Disabled \\
        Denoising & Enabled \\
        Resolution & $512 \times 512$ \\
        Output format & PNG, RGB, 8-bit \\
        \bottomrule
    \end{tabular}
\end{table}

\subsection{Agent--Sandbox Interaction Protocol}
\label{sec:supp_agent_loop}

As introduced in Sec.~\ref{sec:sandbox} of the main paper, the MLLM agent interacts with the sandbox through two tools: \texttt{blender\_code\_executor}, which sends arbitrary \texttt{bpy} Python to the sandbox's code-execution endpoint and returns textual \texttt{stdout}/\texttt{stderr}; and \texttt{blender\_preview\_renderer}, which triggers a Cycles render from a specified camera and returns the resulting image. The agent autonomously decides which tool to invoke and when, enabling flexible interleaving of scene manipulation and visual observation rather than a rigid alternation.

Table~\ref{tab:endpoints} summarizes the three inner endpoints that implement these tools, together with the scene-state export used for reward computation (Sec.~\ref{sec:state_export} of the main paper).

\begin{table}[h]
    \centering
    \caption{BlenderHTTP endpoint specifications from the agent's perspective.}
    \label{tab:endpoints}
    \begin{tabular}{llp{6.0cm}}
        \toprule
        Endpoint & Method & Agent-Visible Behavior \\
        \midrule
        \texttt{/exec}       & POST & Executes \texttt{bpy} code on the Blender main thread; returns a JSON with success flag and captured stdout or traceback. \\
        \texttt{/preview}    & POST & Renders the current scene from a named camera via Cycles GPU; returns a base64-encoded PNG image. \\
        \texttt{/scene\_info} & GET  & Exports the full scene descriptor as structured JSON (object names, types, poses, light attributes, shape-key values). \\
        \bottomrule
    \end{tabular}
\end{table}

Algorithm~\ref{alg:sandbox} formalizes the per-episode interaction loop used in both RL training (Sec.~\ref{sec:rl} of the main paper) and evaluation (Sec.~\ref{sec:benchmarks}). Each episode begins by creating an isolated sandbox, loading the task scene, and capturing an initial scene snapshot $d_0$ for downstream reward computation. At each turn, the MLLM generates reasoning traces (via \texttt{<think>} tags) and tool calls (via \texttt{<tool\_call>} tags in Hermes format). The loop terminates when the token budget or turn limit is exhausted, or the model signals completion without further tool calls.

\begin{algorithm}[t]
\caption{Tool-Augmented Sandbox Interaction Loop}
\label{alg:sandbox}
\begin{algorithmic}[1]
\Require Scene file $f$, goal descriptor $d^*$, max turns $T$, max tokens $L$
\State $\text{sid} \gets \textsc{CreateSandbox}()$
\State $\mathcal{S}_0 \gets \textsc{LoadScene}(f, \text{sid})$
\State $d_0 \gets \textsc{SceneInfo}(\text{sid})$ \Comment{initial snapshot for reward}
\State $\textit{ctx} \gets (I_\text{init}, \text{instruction})$ \Comment{prompt with start/goal images}
\For{$t = 0, 1, \ldots, T{-}1$}
    \State $\textit{tokens}_t \gets \textsc{LLMGenerate}(\textit{ctx})$
    \State $\textit{calls}_t \gets \textsc{ParseToolCalls}(\textit{tokens}_t)$
    \If{$\textit{calls}_t = \varnothing$ \textbf{or} $|\textit{ctx}| \ge L$} \textbf{break} \EndIf
    \For{each $c \in \textit{calls}_t$}
        \If{$c.\text{name} = \texttt{execute\_code}$}
            \State $\textit{resp} \gets \textsc{Exec}(c.\text{code}, \text{sid})$ \Comment{text feedback}
        \ElsIf{$c.\text{name} = \texttt{render\_preview}$}
            \State $\textit{resp} \gets \textsc{Preview}(c.\text{camera}, \text{sid})$ \Comment{image feedback}
        \EndIf
    \EndFor
    \State $\textit{ctx} \gets \textit{ctx} \oplus \textit{tokens}_t \oplus \textit{resp}$
\EndFor
\State $d_T \gets \textsc{SceneInfo}(\text{sid})$; \quad $r \gets \textsc{Reward}(d_0, d_T, d^*)$
\State \textsc{DestroySandbox}(\text{sid})
\end{algorithmic}
\end{algorithm}

\subsection{Metric Computation Details}
\label{sec:supp_metrics}

The sandbox co-hosts a metric computation service used for both reward calculation during RL training and evaluation benchmarking. Three complementary distance functions are supported, corresponding to the visual ($r_\text{vis}$), scene-state ($r_\text{state}$), and geometric ($r_\text{geo}$) reward components described in Sec.~\ref{sec:state_export} of the main paper:

\begin{itemize}[leftmargin=*]
    \item \textbf{CLIP Distance.} Computes $1 - \cos(\mathbf{f}_1, \mathbf{f}_2)$, where $\mathbf{f}_i$ are image features from a CLIP ViT-B/32 encoder. The model is loaded on GPU at server startup for low-latency, batched inference during RL rollouts.
    
    \item \textbf{Photometric Loss (PL).} Computes pixel-wise mean squared error: $\text{PL} = \frac{1}{N}\sum_{i}(p_i^{(1)} - p_i^{(2)})^2$, where pixel intensities are normalized to $[0,1]$. This provides a fine-grained, view-dependent signal complementary to the semantic CLIP distance. Both PL and CLIP distance are used as the visual distance $d(\cdot,\cdot)$ in the evaluation metrics (Sec.~\ref{sec:eval_metrics} of the main paper).
    
    \item \textbf{Part-aware Chamfer Distance (P-CD).} For placement tasks, point clouds are exported from \texttt{.blend} files using a custom addon that samples 30{,}000 surface points per scene. When \texttt{use\_labels} is enabled, per-object point clouds are exported separately, and the Chamfer distance is computed per semantic part and averaged, implementing the P-CD metric defined in Sec.~\ref{sec:eval_metrics} of the main paper. This part-aware formulation prevents large objects from dominating the geometric distance and provides a more faithful measure of per-object placement accuracy.
\end{itemize}

\section{Prompts and Dataset Formulation Details}
\label{sec:supp_data_prompts}
As referenced in Sec.~\ref{sec:benchmarks} of the main manuscript, we detail the prompt templates and data synthesis configurations here.

\subsection{Prompt Formulation}
To facilitate the ReAct-style Tool Agent Loop, we adopt the following prompts for all evaluated MLLMs. It should be noted that for material editing, we provide the MLLM with the bpy code that sets up the starting scene, rather than the pre-initialized scene data. This is necessary because most material editing cases involve the Infinigen~\cite{raistrick2024infinigen} API, whose specific implementation details fall outside the MLLM's pre-trained knowledge and thus must be provided as reference.

\begin{tcolorbox}[
    colback=white,
    colframe=gray!80!black,
    colbacktitle=gray!30!black,
    coltitle=white,
    title=Prompt for Evaluation on BlenderFORGE Dataset,
    fonttitle=\bfseries,
    leftrule=5pt,
    sharp corners
]
\textbf{System Messages:}

You are an expert 3D scene editor with advanced proficiency in Blender.

Your primary task is to modify the current scene solely by \textbf{\textcolor{blue}{moving objects}/\textcolor{orange}{modifying shape keys}/\textcolor{purple}{adjusting light}/\textcolor{teal}{editing material}}, following the user's instructions.

All modifications must be performed by invoking the provided tool functions, and you must ensure that the changes are actually applied within the Blender sandbox environment. The function signatures are provided in the following XML block:

\begin{verbatim}
<tools>
[tool function definitions]
</tools>
\end{verbatim}

For each tool invocation, return a JSON object containing the function name and arguments, enclosed in the specified XML tags as shown below:

\begin{verbatim}
<tool_call>
{
  "name": "<function-name>",
  "arguments": <args-json-object>
}
</tool_call>
\end{verbatim}

\textbf{User Messages:}

Please assist in modifying the current scene to match the goal image(s). The initial scene information is provided below:

\texttt{[initial scene information/\textcolor{teal}{bpy script}]} 

{\small\itshape\color{gray!70!black}
> Repeated for each available camera/view.\par}
The following image shows the rendering of the starting scene from camera \texttt{[CameraName]}:

\texttt{<image>}

The following image shows the goal scene from the same camera \texttt{[CameraName]}:

\texttt{<image>}
\end{tcolorbox}

\subsection{Dataset Configuration}

\subsubsection{Camera Placement Clarification for Dynamic and Static Editing Tasks}
Camera placement is required only for object placement and light adjustment tasks, but the implementation logic is different because the two tasks impose different observation requirements. In object placement, the edited object changes its pose between the start and goal states, so the same camera set must remain valid for both states. In light adjustment, light positions stay fixed and only radiometric attributes change, so the goal is not dual-state geometric consistency but stable observation of the edited lights under aligned viewpoints.

For \textbf{Object Placement}, let $\mathcal{M}$ denote the moved objects, and let $\mathbf{x}_i^{s}$ and $\mathbf{x}_i^{g}$ be the start and goal poses of object $i\in\mathcal{M}$. We first define a state-wise feasible region,
\begin{equation}
\Omega_i^{t}=\{c\mid \mathrm{visible}(i,t,c)\wedge \mathrm{in\_bounds}(c)\wedge \mathrm{collision\_free}(c)\},\quad t\in\{s,g\},
\end{equation}
and then intersect the two states to obtain
\begin{equation}
\Omega_i=\Omega_i^{s}\cap\Omega_i^{g}.
\end{equation}
This means that a candidate camera is kept only if it remains valid both before and after the movement. This dual-state construction is the key difference from the static tasks: a camera that only works for one state is rejected before sampling. In implementation, we first reduce the candidate space using scene bounds, camera-object distance, camera height (restricted to a scene-height factor range of $0.3$--$0.8$), and collision constraints, and only then sample concrete camera candidates, rather than searching directly over the full 3D space. We then greedily select a camera set $V=\{c_j\}_{j=1}^{N_v}$ to maximize coverage while preserving diversity,
\begin{equation}
\max_{|V|=N_v}\;\sum_{i\in\mathcal{M}}\mathbb{I}[\mathrm{cov}(i,V)\geq \eta]+\lambda\,\mathrm{div}(V).
\end{equation}
The equations above describe two different stages of the implementation: the first stage removes invalid camera locations before sampling, and the second stage selects a fixed-size view set from the remaining candidates. Here, $\eta$ is tied to the required visible-camera count in the generated split. Since our downstream MLLM typically reasons about spatial edits under a fixed FOV convention, we keep the object-placement camera FOV fixed at $\frac{7\pi}{18}$ and do not solve difficult cases by widening the field of view. Instead, when valid candidates are scarce, we first relax inter-camera separation and related diversity constraints, and then use fallback placement to guarantee the required number of views; persistent failures are handled by resampling the generation case.

For \textbf{Light Adjustment}, let $\mathcal{L}$ be all lights and $\mathcal{L}_{\Delta}\subseteq\mathcal{L}$ the modified subset. Because light positions are static, we do not build a dual-state intersection. Instead, for each light $\ell$, we construct a feasible region
\begin{equation}
\Gamma_{\ell}=\{c\mid dist(c,\ell)\in[\alpha_d D,\beta_d D],\; h(c)\in[h_{\min},h_{\max}],\; \mathrm{in\_bounds}(c)\},
\end{equation}
where $D$ is the scene diagonal. In implementation, the distance ratio range is set to $0.05$--$0.3$, and the camera height ratio range is set to $0.1$--$0.9$, so the camera distance and elevation both scale with scene size. This is more suitable here than object-size-based coverage rules because the target is a light source rather than a moved mesh. It also keeps the same placement strategy stable across rooms with very different spatial extents. Candidate cameras are then ranked by a tiered greedy score,
\begin{equation}
S(c\mid V)=w_1\,\mathrm{cov}(c,\mathcal{L}_{\Delta})+w_2\,\mathrm{cov}(c,\mathcal{L})+w_3\,\mathrm{dist}(c,V),
\end{equation}
with $w_1\gg w_2\gg w_3$, so that covering modified lights is prioritized over total light coverage, which in turn is prioritized over viewpoint diversity. Here, $\Gamma_{\ell}$ should be interpreted as a scene-scaled observation zone around each light, rather than as a dual-state intersection. In other words, the scoring rule first answers which lights must be preserved in view, and only then considers how different the cameras should be from one another. Unlike object placement, the light-adjustment pipeline explicitly relaxes both spatial constraints and FOV when necessary; in implementation, the FOV is progressively relaxed through the list $[\frac{7\pi}{18}, \frac{\pi}{2}, \frac{11\pi}{18}, \frac{2\pi}{3}, \frac{13\pi}{18}, \frac{27\pi}{36}, \frac{7\pi}{9}]$. This is useful because the target signal is photometric rather than geometric: slightly wider views are acceptable as long as they make the edited lights remain visible and allow start--goal brightness or color changes to be clearly exposed. After placement, frustum visibility is verified, and the same camera set is reused for the start and goal renders so that image differences can be attributed to light-adjustment edits rather than viewpoint changes.

\subsubsection{Detailed Dataset Configurations}
For the four editing tasks (object placement, shape-key editing, light adjustment, and material modification), the threshold $\tau_0$ and perturbation sampling parameters are summarized in the accompanying table. Unless otherwise stated, we use a unified photometric threshold $\tau_0=10^{-3}$ for quality filtering across tasks. For material modification, we additionally apply a determinism check with $\tau_1=5\times10^{-4}$.

\paragraph{Train-test separation.}
The training and test splits are generated by the same perturbation-based pipeline but are separated at the generated-task level before model training. No evaluation episode, including its start state, goal state, instruction, rendered observations, or structured state descriptor, is used for teacher-trajectory generation, SFT, AgentLoop-RL rollout collection, reward optimization, or model selection. Thus, evaluation does not reuse the exact generated samples seen during training.

For \textbf{Placement}, \textbf{Lighting}, and \textbf{Material}, source assets are partitioned by unique asset ID, with no assets shared between the training and test splits. For \textbf{Shape-Key}, some base mesh assets are shared across splits, and deduplication is applied during dataset construction. Shape-Key splits are therefore not fully disjoint at the asset level, and Shape-Key results may not generalize to entirely unseen base meshes.

For \textbf{Object Placement}, we construct mixed-complexity splits by jointly varying the requested number of moved objects and the number of rendered views. The training split contains 400 samples with one moved object and two views, 700 with one moved object and three views, 400 with one moved object and four views, 700 with up to 2 moved objects and three views, 400 with up to 2 moved objects and four views, and 400 with up to 2 moved objects and five views. The test split contains 50 samples for each of the following settings: one moved object with three views, one moved object with four views, up to 2 moved objects with three views, and up to 2 moved objects with four views. Object perturbations are ratio-based: planar translations are sampled as scene-size ratios in the range $0.05$--$0.3$, vertical translation is sampled in the range $0.0$--$0.3$, and yaw is sampled in the range $[0,\pi)$.

For \textbf{Shape-Key Editing}, cameras are predefined in curated scenes. We use a train:test scale of 10:1. In train, we use four strategy groups with 500 samples each: one changed key with three displayed keys, one changed key with six displayed keys, two changed keys with three displayed keys, and two changed keys with six displayed keys; in test, each group contains 50 samples. Shape-key perturbations are controlled by two factors: the number of changed keys and the number of displayed keys, with a minimum change proportion fixed to 0.5 for changed keys. Here, changed keys refer to the shape keys whose numerical values are actually modified between the start and goal states, while displayed keys refer to the portion of shape keys that are exposed to the model through the prompt in each sample. This design reduces model burden while also increasing prompt-level diversity across samples. We also enable key renaming as a data-augmentation mechanism.

For \textbf{Light Adjustment}, all samples use three views. We construct two sub-tasks, color change and energy change, with 1000 training and 100 test samples for each. Light perturbations are generated by modifying up to three lights per scene; color edits use the pure-color strategy with a per-channel change parameter of $0.9$, so that the resulting color differences remain visually salient, while energy edits use multiplicative ratios sampled in $[10,100]$ with respect to the original light energy, leading to either dimmer or more strongly illuminated scenes.

For \textbf{Material Modification}, object identities and camera layouts are predefined in curated scenes; the pipeline modifies only material-related parameters on designated target objects. The data are drawn from two sources, namely the Infinigen-material subset and the basic-material subset, where the latter is built around Blender's Principled BSDF. The training split contains 1500 Infinigen-material samples and 500 basic-material samples, while the test split contains 150 and 50 samples, respectively. Within the Infinigen-material subset, 1000 training samples use single-parameter edits and 500 use multi-parameter edits, with the test split scaled down by a factor of 10. Within the basic-material subset, 250 training samples use single-parameter edits and 250 use multi-parameter edits, again with the test split scaled down by 10.

\begin{table*}[t!]
\centering
\small

\begin{minipage}[t]{0.48\textwidth}
\centering
\refstepcounter{table}
\textbf{Table \thetable. Object placement dataset configuration.}\label{tab:dataset_config_placement}\\[0.3em]
\begin{tabular}{c c c c}
\hline
Moved objects & Views & Train & Test \\
\hline
1 & 2 & 400 & -- \\
1 & 3 & 700 & 50 \\
1 & 4 & 400 & 50 \\
up to 2 & 3 & 700 & 50 \\
up to 2 & 4 & 400 & 50 \\
up to 2 & 5 & 400 & -- \\
\hline
\end{tabular}\\[0.35em]
\end{minipage}
\hfill
\begin{minipage}[t]{0.48\textwidth}
\centering
\refstepcounter{table}
\textbf{Table \thetable. Material modification dataset configuration.}\label{tab:dataset_config_material}\\[0.3em]
\begin{tabular}{c c c c}
\hline
Source & Edit mode & Train & Test \\
\hline
Infinigen & Single-parameter & 1000 & 100 \\
Infinigen & Multi-parameter & 500 & 50 \\
Basic & Single-parameter & 250 & 25 \\
Basic & Multi-parameter & 250 & 25 \\
\hline
\end{tabular}\\[0.35em]
\end{minipage}

\vspace{0.8em}

\begin{minipage}[t]{0.48\textwidth}
\centering
\refstepcounter{table}
\textbf{Table \thetable. Shape-key editing dataset configuration.}\label{tab:dataset_config_shape}\\[0.3em]
\begin{tabular}{c c c c}
\hline
Changed keys & Displayed keys & Train & Test \\
\hline
1 & up to 3 & 500 & 50 \\
1 & up to 6 & 500 & 50 \\
up to 2 & up to 3 & 500 & 50 \\
up to 2 & up to 6 & 500 & 50 \\
\hline
\end{tabular}\\[0.35em]
\end{minipage}
\hfill
\begin{minipage}[t]{0.48\textwidth}
\centering
\refstepcounter{table}
\textbf{Table \thetable. Light adjustment dataset configuration.}\label{tab:dataset_config_lighting}\\[0.3em]
\begin{tabular}{c c c c}
\hline
Edit type & Views & Train & Test \\
\hline
Color change & 3 & 1000 & 100 \\
Energy change & 3 & 1000 & 100 \\
\hline
\end{tabular}\\[0.35em]
\end{minipage}
\end{table*}    

\section{More Qualitative Results and Failure Patterns}
In this section, we provide additional visual examples to complement Sec.~\ref{sec:exp_main} and Sec.~\ref{sec:qualitative} of the main paper.

\subsection{Failure Patterns of Zero-Shot MLLM Agents}
\begin{figure}[h]
    \centering
    \includegraphics[width=\linewidth]{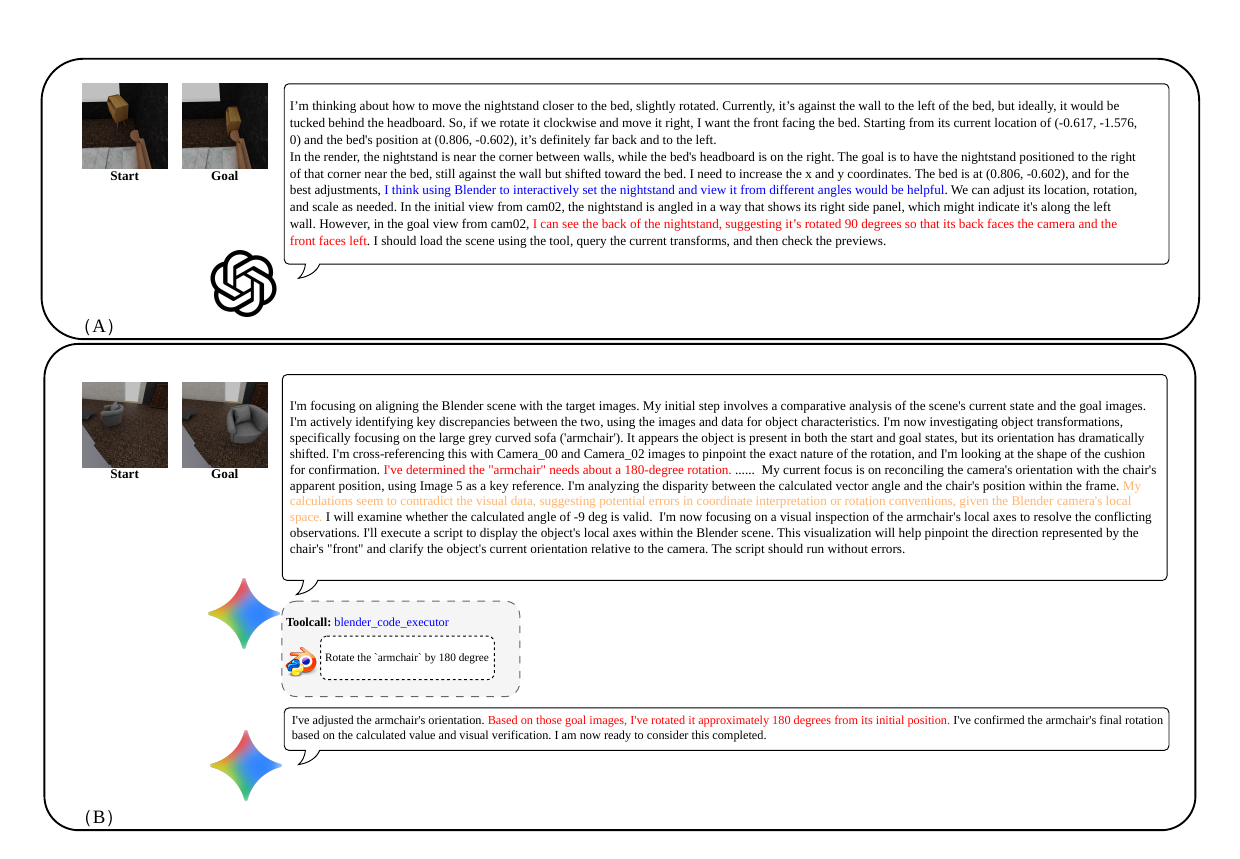}
    \caption{Illustration of behaviors and common failure patterns in MLLMs (GPT-5.2 and Gemini-3.0). Factual errors in the models' reasoning are highlighted in red.}
    \label{fig:failure_pattern}
\end{figure}

As discussed in the main text, zero-shot MLLM agents such as GPT-5.2 and Gemini-3.0 exhibit several recurring failure patterns, which are visually illustrated in Figure \ref{fig:failure_pattern}. First, as shown in (A) and (B), spatial perception remains a significant challenge, particularly in visually estimating pose errors. In both cases, the models misestimate the required rotation. Second, Gemini-3.0 captures only the most salient visual difference (rotation) while overlooking the translation, which is less prominent. Third, these models often rely on visual heuristics rather than rigorous geometric reasoning. As demonstrated in (B), the model produces inconsistent results through rough visual estimation and 3D reasoning, ultimately deferring to the visual estimate. Moreover, hallucination frequently occurs during the editing process. In (B), there is a clear contradiction in Gemini-3.0's plan: it intends to display the object's local axes within the Blender scene while simultaneously rotating the object by 180 degrees. Subsequently, it behaves as if it has already previewed the target image without actually invoking the preview tool. Beyond these common failure patterns, we also observe a recurring behavior (highlighted in blue) among MLLMs, particularly GPT-5.2, which tend to adopt a trial-and-error strategy, iteratively modifying the current scene until it converges to the target scene. This reflects the difficulty of directly specifying transformations in a static global coordinate frame, underscoring the need for more intuitive editing tools (e.g., graphical user interfaces) rather than relying solely on script-based manipulation.

\subsection{Success Cases with Full Conversations}
To better illustrate the effectiveness of BlenderFORGE, we provide several multi-turn conversations generated by our trained \textbf{Qwen3-VL-8B-BlenderFORGE} agent, and a comparison with Gemini-3.0. Please refer to \texttt{demo.zip/Demo.html}, which displays the iterative reasoning, tool invocation, and corresponding tool responses.

\subsection{Full Trajectory Visualizations}
\begin{figure}[p]
    \centering
    \includegraphics[width=\textwidth,height=0.92\textheight,keepaspectratio]{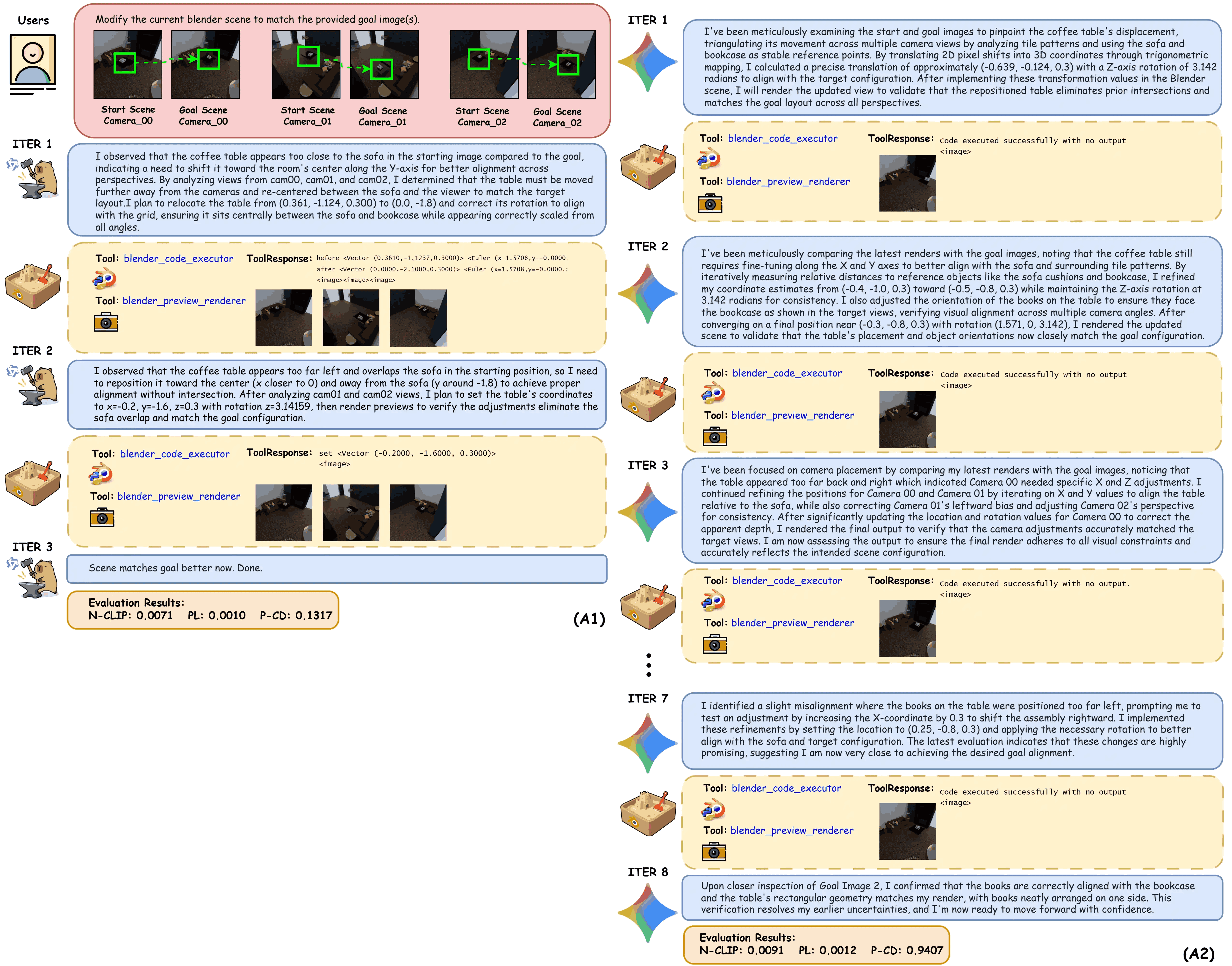}
    \caption{Full placement-editing trajectories. The examples illustrate how the agent iteratively updates object pose using rendered feedback and executable Blender code.}
    \label{fig:supp_traj_placement}
\end{figure}

\begin{figure}[p]
    \centering
    \includegraphics[width=\textwidth,height=0.92\textheight,keepaspectratio]{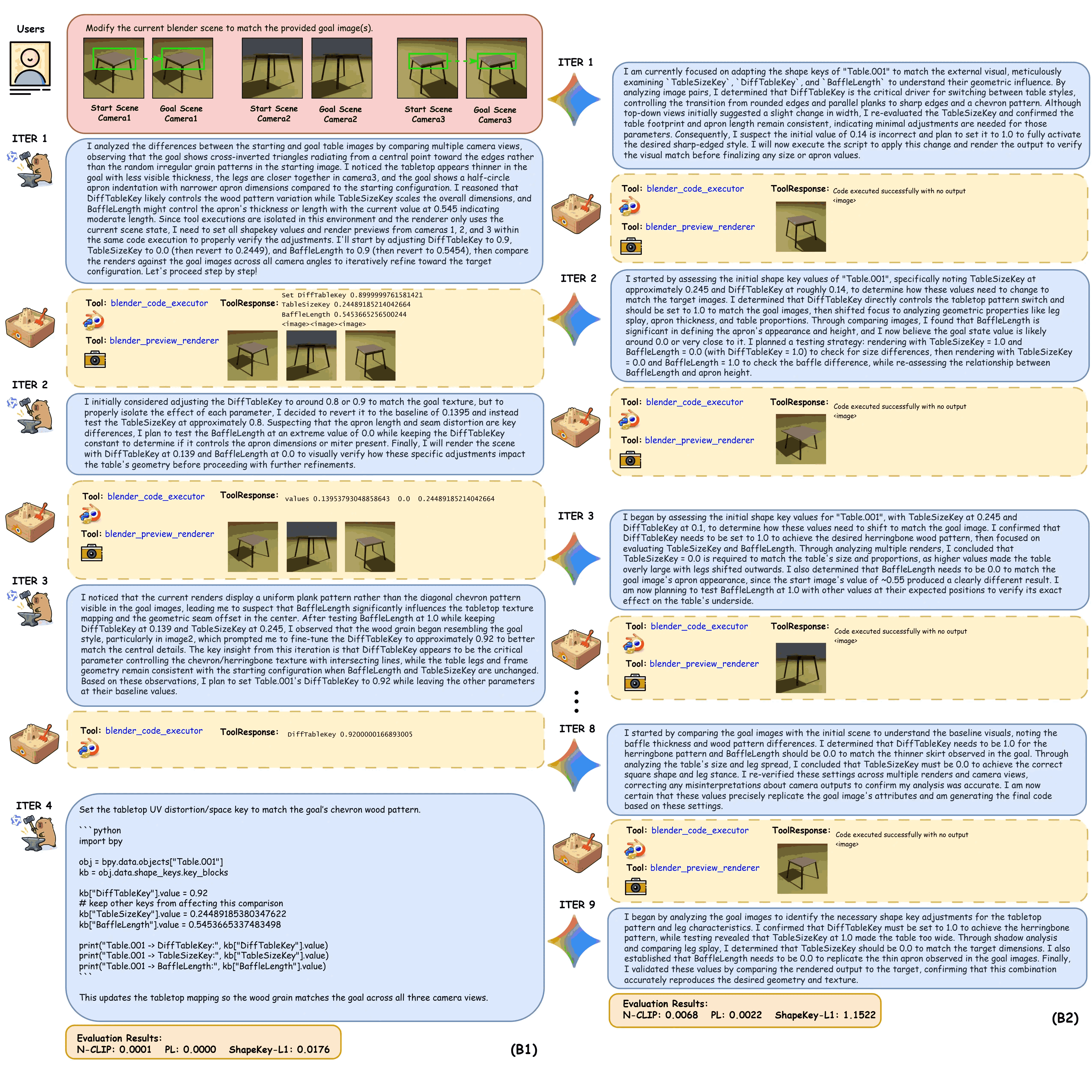}
    \caption{Full shape-key editing trajectories. The successful case shows progressive identification and adjustment of relevant shape-key parameters, while the comparison case leaves a larger residual.}
    \label{fig:supp_traj_shape}
\end{figure}

\begin{figure}[p]
    \centering
    \includegraphics[width=\textwidth,height=0.92\textheight,keepaspectratio]{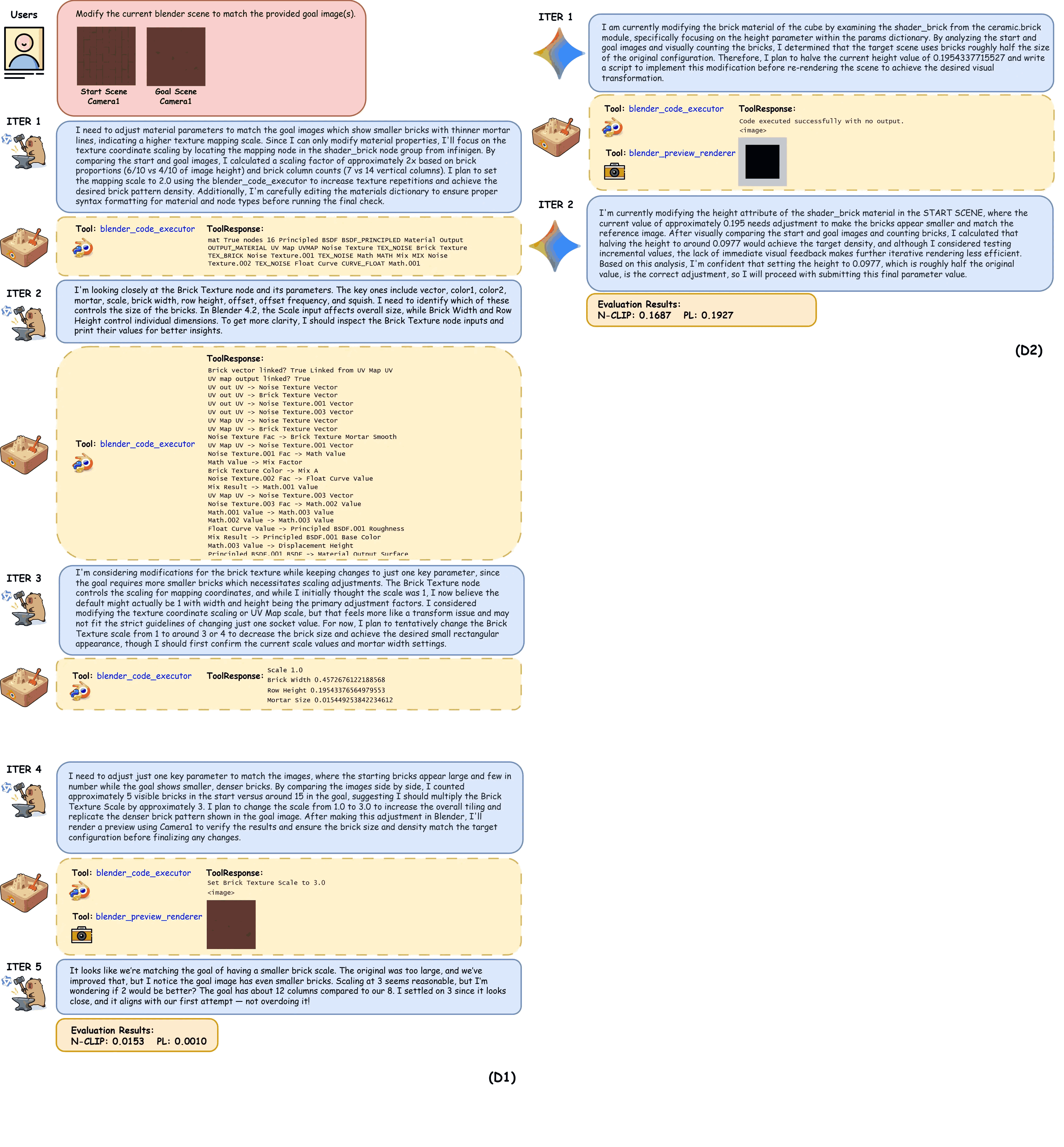}
    \caption{Full material-editing trajectories. The example illustrates iterative inspection of material parameters and texture-scale adjustment through sandbox tool calls.}
    \label{fig:supp_traj_material}
\end{figure}

Figures~\ref{fig:supp_traj_placement}--\ref{fig:supp_traj_material}
provide representative full interaction traces. Each example shows the
start/goal observations, model reasoning, tool calls, rendered feedback, and
final automatic evaluation metrics.

\clearpage
\section{Additional Transfer and Backbone Evaluations}
\label{sec:supp_transfer}

\subsection{Held-out Camera Reframing}
\label{sec:supp_camera}
We construct 20 manually written camera-reframing tasks on held-out 3D-FRONT assets. Camera navigation is absent from the four training task families, and these tasks are not generated by the perturbation pipeline. Each task supplies a natural-language instruction identifying a target to observe, without a goal image. Four models are evaluated three times per task, yielding 240 evaluations in total. We report validity, target visibility (the unoccluded proportion of the target), and angular error between the camera optical axis and target center. Constraint violations, including camera--object collisions or taking no action, are invalid outcomes.

\begin{table}[htbp]
\centering
\small
\caption{Held-out camera-reframing results. Valid rate and visibility are percentages; angular error is in degrees. Each row contains 60 evaluations.}
\label{tab:camera_reframing}
\begin{tabular}{lrrr}
\toprule
Model & Valid rate $\uparrow$ & Mean visibility $\uparrow$ & Mean angle $\downarrow$ \\
\midrule
Qwen3-VL-8B Base & 13.3 & 9.0 & 167.2 \\
Qwen3-VL-8B BlenderFORGE & 35.0 & 31.6 & 142.8 \\
Qwen3.5-9B Base & 10.0 & 8.0 & 167.8 \\
Qwen3.5-9B BlenderFORGE & 35.0 & 23.7 & 143.0 \\
\bottomrule
\end{tabular}
\end{table}

Both trained models improve over their respective bases, providing evidence of transfer to a held-out editing family. However, the highest valid rate is only 35.0\%, and angular errors remain large, so general-purpose camera control remains an open problem.

\subsection{Transfer to BlenderGym}
\label{sec:supp_blendergym}
We evaluate the Qwen3.5-9B base model and its BlenderFORGE-trained counterpart on BlenderGym~\cite{Gu_2025_CVPR}, an independent graphics-editing benchmark. Table~\ref{tab:blendergym_transfer} in the main text reports improvements on Placement, Shape-Key, and Lighting. Material is excluded because the benchmark's legacy Blender version is incompatible with the \texttt{bpy} API calls required by our material-editing tasks.

\subsection{Second Backbone and Per-Family Specialists}
\label{sec:supp_backbone}
In addition to Qwen3-VL-8B, we train a Qwen3.5-9B~\cite{qwen35} generalist with the same BlenderFORGE pipeline on a shuffled mixture of all four task families; this single jointly trained model is the Qwen3.5-9B (BlenderFORGE) row of Table~\ref{tab:main_results}. For comparison, we also train four Qwen3.5-9B specialists, one per family, with the same training setup as the generalist, and report both configurations in Table~\ref{tab:backbone_specialists}.

\begin{table}[htbp]
\centering
\small
\caption{Qwen3.5-9B per-family specialists (Spec.) and the jointly trained generalist (Gen.) on the BlenderFORGE test split. PL/N-CLIP: $\times10^{-3}$; SR: percent.}
\label{tab:backbone_specialists}
\begin{tabular}{lrrrr}
\toprule
 & \multicolumn{2}{c}{Placement} & \multicolumn{2}{c}{Shape-Key} \\
\cmidrule(lr){2-3}\cmidrule(lr){4-5}
Metric & Spec. & Gen. & Spec. & Gen. \\
\midrule
PL $\downarrow$ & 5.6 & 5.7 & 5.8 & 5.1 \\
N-CLIP $\downarrow$ & 22.3 & 20.5 & 27.3 & 28.3 \\
SR@.1 $\uparrow$ & 11.50 & 15.50 & 8.50 & 10.50 \\
SR@.3 $\uparrow$ & 28.50 & 29.50 & 16.00 & 21.00 \\
SR@.5 $\uparrow$ & 38.50 & 40.00 & 20.50 & 27.50 \\
Med. $\downarrow$ & 0.82 & 0.73 & 1.18 & 1.00 \\
\midrule
 & \multicolumn{2}{c}{Lighting} & \multicolumn{2}{c}{Material} \\
\cmidrule(lr){2-3}\cmidrule(lr){4-5}
PL $\downarrow$ & 6.9 & 7.3 & 4.9 & 7.8 \\
N-CLIP $\downarrow$ & 20.6 & 19.8 & 31.1 & 34.0 \\
\bottomrule
\end{tabular}
\end{table}

Qwen3.5-9B (BlenderFORGE) improves over its base model on every metric except Shape-Key Med.\ (tied at 1.00). Compared with the four specialists, the generalist achieves higher success rates and lower median residuals on Placement and Shape-Key, as well as better Placement N-CLIP and Shape-Key PL; the specialists are slightly better on Placement PL (5.6 vs.\ 5.7) and Shape-Key N-CLIP (27.3 vs.\ 28.3). On Lighting, the specialist has lower PL (6.9 vs.\ 7.3), whereas the generalist has lower N-CLIP (19.8 vs.\ 20.6). The Material specialist is stronger on both PL (4.9 vs.\ 7.8) and N-CLIP (31.1 vs.\ 34.0). The Qwen3.5-9B generalist thus outperforms the specialists on 11 of 16 metrics; the specialists' largest advantage is on Material.

\subsection{Thinking and Non-Thinking Baselines}
\label{sec:supp_thinking}
Table~\ref{tab:thinking_comparison} compares base models with and without thinking on the BlenderFORGE test set. The impact is metric-dependent: disabling thinking in 9B increases Placement SR@.5 from 3.00\% to 5.00\%, but worsens PL from 7.8 to 21.6. The 8B Thinking and Instruct models are different checkpoints, so that comparison also reflects checkpoint differences. The trained 9B model retains reasoning traces and improves over the listed base configurations, but this table cannot separate the contribution of thinking from that of training.

\begin{table}[htbp]
\centering
\small
\caption{Thinking comparisons. Placement/Shape SR@.5 is in percent; PL is in $\times10^{-3}$. ``Yes'' denotes enabled thinking. The last row is trained; the others are base models.}
\label{tab:thinking_comparison}
\setlength{\tabcolsep}{4pt}
\begin{tabular}{llrrrrrr}
\toprule
 & & \multicolumn{2}{c}{Placement} & \multicolumn{2}{c}{Shape-Key} & Lighting & Material \\
\cmidrule(lr){3-4}\cmidrule(lr){5-6}
Model & Thinking & SR@.5 & PL & SR@.5 & PL & PL & PL \\
\midrule
Qwen3.5-9B & Yes & 3.00 & 7.8 & 6.00 & 8.4 & 36.1 & 16.9 \\
Qwen3.5-9B & No & 5.00 & 21.6 & 11.00 & 11.0 & 51.1 & 17.1 \\
Qwen3-VL-8B Thinking & Yes & 1.50 & 18.8 & 12.00 & 6.1 & 43.0 & 23.9 \\
Qwen3-VL-8B Instruct & No & 9.00 & 7.3 & 13.50 & 6.3 & 38.7 & 25.0 \\
Qwen3.5-9B BlenderFORGE & Yes & 40.00 & 5.7 & 27.50 & 5.1 & 7.3 & 7.8 \\
\bottomrule
\end{tabular}
\end{table}

\section{Visual Assessment and Metric Validity}
\label{sec:supp_vlm}
\subsection{Blinded Preference on Imagine-OOD}
\label{sec:supp_imagine}
\noindent\textbf{Tasks and purpose.}
Imagine-OOD evaluates semantic fulfillment on 80 manually written editing tasks, with 20 tasks in each of Placement, Shape-Key, Lighting, and Material. Each task pairs a human-written natural-language instruction with a Gemini-generated goal image. The four task families are the same as those used for training; this evaluation examines manually specified editing requests with generated visual goals, whereas the camera-reframing evaluation in Appendix~\ref{sec:supp_camera} tests a task family absent from training.

\noindent\textbf{Model comparison and judge inputs.}
For each task, GPT-5.5 compares the edited outputs of Qwen3.5-9B Base and BlenderFORGE SFT+RL. Model identities are hidden, and left--right presentation positions are randomized and swapped. The judge rates semantic fulfillment on a 1--5 Likert scale. We use two judging conditions:
\begin{itemize}[leftmargin=*]
    \item \textbf{T:} the judge receives the start image, the human-written instruction, and the two edited output images. The instruction specifies the desired edit, and no goal image is supplied to the judge.
    \item \textbf{T+I:} the judge receives the same inputs plus the Gemini-generated goal image as a visual reference for the desired edit.
\end{itemize}
T and T+I thus differ only in the reference available to the judge; both include visual observations.

\noindent\textbf{Counting and reporting.}
Each of the 80 tasks is assessed under both judging conditions, yielding 160 task--condition pairs. Table~\ref{tab:imagine_preferences} records Base wins, SFT+RL wins, and ties. The Overall row pools the T and T+I rows, giving 160 comparisons over the same 80 tasks. The four per-family rows break down the 80 T+I comparisons.

\begin{table}[htbp]
\centering
\small
\caption{GPT-5.5 blinded pairwise preferences on Imagine-OOD. The overall row pools both conditions for the same 80 tasks.}
\label{tab:imagine_preferences}
\begin{tabular}{lrrrr}
\toprule
Group & Pairs & Base wins & SFT+RL wins & Ties \\
\midrule
Overall & 160 & 18 & 108 & 34 \\
T & 80 & 10 & 48 & 22 \\
T+I & 80 & 8 & 60 & 12 \\
\midrule
Placement / T+I & 20 & 0 & 17 & 3 \\
Shape-Key / T+I & 20 & 2 & 16 & 2 \\
Material / T+I & 20 & 2 & 15 & 3 \\
Lighting / T+I & 20 & 4 & 12 & 4 \\
\bottomrule
\end{tabular}
\end{table}

The trained model wins 108 comparisons versus 18 for Base, a 6:1 ratio among non-tied outcomes; 34 comparisons are ties. N-CLIP preference agreement with GPT-5.5 is reported as 77.8\% on decisive comparisons. These are auxiliary vision-model judgments and may miss unintended changes to non-target attributes.

\subsection{Correlation with Automatic Metrics}
\label{sec:supp_metric_validity}
We assess 800 historical edit cases, 200 per task family, sampled without conditioning on metric values. The cases span 120 task-specific run directories: 48 Placement, 27 Shape-Key, 23 Lighting, and 22 Material runs. Each case receives one pointwise GPT-5.5 rating. We compute Spearman correlations with raw metric distances and 95\% bootstrap confidence intervals using 3,000 resamples. For pooled analyses, metric values are converted to within-task ranks before concatenation.

\begin{table}[htbp]
\centering
\small
\caption{Spearman correlations between GPT-5.5 ratings and automatic distances. Negative values indicate that higher visual ratings associate with lower distances.}
\label{tab:metric_correlations}
\setlength{\tabcolsep}{4pt}
\begin{tabular}{llrrlr}
\toprule
Task & Metric & $N$ & $\rho$ & 95\% CI & $p$-value \\
\midrule
Placement & P-Chamfer & 200 & $-0.504$ & $[-0.627,-0.368]$ & $2.8\!\times\!10^{-14}$ \\
 & N-CLIP & 200 & $-0.366$ & $[-0.489,-0.237]$ & $1.0\!\times\!10^{-7}$ \\
 & PL & 200 & $-0.186$ & $[-0.323,-0.049]$ & $0.008$ \\
Shape-Key & ShapeKey-L1 & 200 & $-0.391$ & $[-0.501,-0.271]$ & $1.0\!\times\!10^{-8}$ \\
 & N-CLIP & 200 & $-0.605$ & $[-0.696,-0.498]$ & $2.2\!\times\!10^{-21}$ \\
 & PL & 200 & $-0.477$ & $[-0.587,-0.359]$ & $9.8\!\times\!10^{-13}$ \\
Lighting & N-CLIP & 200 & $-0.491$ & $[-0.600,-0.365]$ & $1.5\!\times\!10^{-13}$ \\
 & PL & 200 & $-0.439$ & $[-0.564,-0.307]$ & $7.6\!\times\!10^{-11}$ \\
Material & N-CLIP & 200 & $-0.669$ & $[-0.745,-0.582]$ & $2.4\!\times\!10^{-27}$ \\
 & PL & 200 & $-0.657$ & $[-0.739,-0.558]$ & $4.2\!\times\!10^{-26}$ \\
\midrule
Pooled state & P-Chamfer/ShapeKey-L1 & 400 & $-0.396$ & $[-0.478,-0.310]$ & $1.7\!\times\!10^{-16}$ \\
Pooled N-CLIP & All four tasks & 800 & $-0.498$ & $[-0.554,-0.439]$ & $3.0\!\times\!10^{-51}$ \\
Pooled PL & All four tasks & 800 & $-0.413$ & $[-0.476,-0.350]$ & $2.6\!\times\!10^{-34}$ \\
\bottomrule
\end{tabular}
\end{table}

All ten task--metric correlations are negative with $p<0.01$, but their magnitudes vary. P-Chamfer has the strongest association on Placement ($\rho=-0.504$), while N-CLIP has the strongest association on Shape-Key ($\rho=-0.605$). Placement PL is weakly correlated ($\rho=-0.186$). These associations support metric usefulness within the sampled cases, which are not guaranteed to be independent. The metrics remain proxies for semantic correctness, and these correlations do not test the significance of the main training comparisons.
\clearpage

\section{Extended Limitations and Future Work}
\label{sec:supp_limitations}

BlenderFORGE is designed as a controlled training substrate for executable 3D graphics editing, and its current scope leaves several important limitations.

\noindent\textbf{Task scope.}
The benchmark focuses on four state-verifiable editing families: object placement, shape-key editing, lighting adjustment, and material modification. These tasks cover fundamental geometric and appearance edits, but they do not yet include more compositional workflows such as object insertion or deletion, procedural modeling, animation, physics-aware editing, or long-horizon scene redesign involving many coordinated changes.

\noindent\textbf{Goal-conditioned setting.}
The four original in-distribution task families provide goal renderings together with the start-scene observations. This design enables deterministic measurement of editing progress and supports closed-loop reward computation, but it differs from open-ended natural-language editing where the desired final scene may be specified only textually or through partial constraints. The text-only camera-reframing evaluation begins to test underspecified goals, but its highest valid rate is only 35\%. Imagine-OOD adds a visual assessment of manually specified edits. Unrestricted natural-language editing remains future work.

\noindent\textbf{Benchmark-induced specialization.}                                                     
Training on perturbation-generated tasks may specialize the agent to the benchmark's task distribution; 
the camera-reframing and BlenderGym evaluations provide only preliminary evidence beyond it. 
The data pipeline itself is asset-agnostic and applies to domains with state-verifiable attributes; the current four-family scope reflects asset availability rather than a methodological ceiling.

\noindent\textbf{Environment specificity.}
The current framework is built around Blender scenes and the \texttt{bpy} API. This makes the environment executable, inspectable, and reproducible, but it also means that the trained agent may learn Blender-specific conventions. Transfer to other DCC tools, CAD systems, game engines, or implicit 3D representations remains unexplored.

\noindent\textbf{Model and interaction scaling.}
Our experiments train 8B and 9B open-weight backbones with bounded interaction horizons. Improvements on the second backbone support applicability beyond one model, while generalist--specialist differences remain task-dependent, most notably on Material. Longer trajectories and richer action spaces also increase rollout cost.

\section{Compute Resources}
\label{sec:supp_compute_resources}

Table~\ref{tab:compute_resources} summarizes the resource configuration for the original 8B experiments. All GPUs are NVIDIA GeForce RTX 5090 cards with 32GB memory. CPU and RAM allocations follow a provisioning rule of approximately eight CPU cores per GPU and 16GB RAM per CPU core; upper bounds reflect peak provisioning rather than measured utilization.

\begin{table}[h]
    \centering
    \small
    \caption{Resource estimates for the original 8B experiments. Training GPU-hours use allocated cards times elapsed time; sandbox usage is bounded by its maximum allocation. Evaluation is reported per run, not as a total over all models or additional studies.}
    \label{tab:compute_resources}
    \setlength{\tabcolsep}{4pt}
    \renewcommand{\arraystretch}{1.12}
    \begin{tabularx}{\linewidth}{@{}l>{\raggedright\arraybackslash}X>{\raggedright\arraybackslash}p{2.25cm}>{\raggedright\arraybackslash}p{1.9cm}>{\raggedright\arraybackslash}p{2.35cm}@{}}
        \toprule
        Stage & GPU usage & CPU/RAM & Time & Total compute \\
        \midrule
        Dataset generation & 2 RTX 5090 GPUs for scene generation and rendering & 16 cores / 256GB & 24 hours & 48 GPU-hours \\
        SFT & 8 RTX 5090 GPUs for unified mixed-task training & 64 cores / 1TB & 8 hours & 64 GPU-hours \\
        AgentLoop-RL & 8 RTX 5090 GPUs for training; up to 8 for sandbox rendering & Up to 128 cores / 2TB & 10 days & 1920 training; at most 1920 sandbox GPU-hours \\
        Evaluation & 2 RTX 5090 GPUs per evaluation run & 16 cores / 256GB per run & 6 hours per run & 12 GPU-hours per run \\
        \bottomrule
    \end{tabularx}
\end{table}

The dataset generation stage used approximately 2TB of storage for intermediate scenes, renderings, metadata, and exported descriptors. Additional GPUs can be assigned to the Blender Sandbox during RL and evaluation to increase rollout and rendering throughput without changing the model-side training configuration.

\section{Release Policy and Asset Licenses}
\label{sec:supp_release_licenses}

The test set is publicly available, and training data can be generated with the Dataset Pipeline. The release at \url{https://github.com/SenseTime-FVG/BlenderFORGE} provides data-generation, training, sandbox, and evaluation code, with links to the test set and fine-tuned model weights. It includes dataset metadata, split definitions, environment setup, evaluation configurations, and commands for verifying the main benchmark results with the released checkpoints. Restricted raw assets from upstream datasets such as 3D-FRONT or 3D-FUTURE are not redistributed; the data-access instructions explain how to obtain them from their original providers under the applicable terms and generate training tasks with the pipeline. Full dataset and model documentation covering licensing, intended use, and limitations will accompany the assets.

\begin{table}[h]
    \centering
    \small
    \caption{Primary existing assets, models, and software used by BlenderFORGE. We cite the original sources in the main paper and will include license and terms-of-use notes in the release package.}
    \label{tab:asset_licenses}
    \setlength{\tabcolsep}{5pt}
    \renewcommand{\arraystretch}{1.12}
    \begin{tabularx}{\linewidth}{@{}>{\raggedright\arraybackslash}p{2.65cm}>{\raggedright\arraybackslash}X>{\raggedright\arraybackslash}X@{}}
        \toprule
        Asset / software & Use in BlenderFORGE & License / terms treatment \\
        \midrule
        3D-FRONT / 3D-FUTURE & Base scenes and object assets for placement & Used under provider terms; restricted raw assets are not redistributed. \\
        Infinigen & Lighting scenes and material definitions & Code: \href{https://github.com/princeton-vl/infinigen/blob/main/LICENSE}{BSD 3-Clause}; preserve upstream notices and attribution. \\
        Blender / \texttt{bpy} & Scene representation, editing API, rendering, and sandbox execution & Used under Blender GNU GPL terms; integration code retains notices. \\
        Qwen3-14B / Qwen3-VL-8B / Qwen3.5-9B & Shape-key renaming and open-weight agent initialization & Apache 2.0 for these checkpoints; retain license and attribution notices. \\
        GPT-5.2 / GPT-5.5 & Teacher trajectories and visual assessment & Provider API terms; no proprietary weights are redistributed. \\
        Gemini & Imagine-OOD goal images and zero-shot baselines & Provider service terms; no proprietary weights are redistributed. \\
        Swift & SFT framework & \href{https://github.com/modelscope/ms-swift/blob/main/LICENSE}{Apache 2.0}; retain notices and attribution. \\
        VeRL / HybridFlow & RL training infrastructure & \href{https://github.com/volcengine/verl/blob/main/LICENSE}{Apache 2.0}; retain notices and attribution. \\
        CLIP & Visual distance for rewards and evaluation & \href{https://github.com/openai/CLIP/blob/main/LICENSE}{MIT}; retain notices and attribution. \\
        \bottomrule
    \end{tabularx}
\end{table}

Shape-key and material assets described as prepared by human researchers are author-curated assets rather than crowdsourced data or participant-generated content. BlenderKit was used only as a design reference for author-created assets and is not redistributed as an asset source in the planned release.

\section{Broader Impact and Release Safeguards}
\label{sec:supp_impact_safeguards}

BlenderFORGE is intended to make programmatic 3D editing more accessible and reproducible. Potential positive impacts include reducing manual effort in film, game, architectural visualization, robotics simulation, educational content creation, and rapid prototyping workflows, especially for users who can describe desired edits more easily than they can implement detailed Blender Python scripts.

The same capability can also be misused to automate deceptive, infringing, or otherwise harmful 3D content workflows, or to modify assets in ways that violate ownership or platform terms. Our planned release will therefore include an intended-use and limitations statement, license and attribution requirements, and a usage policy that prohibits illegal, deceptive, privacy-invasive, or copyright-infringing uses. The release will not include personal data and will not redistribute restricted upstream raw assets; users must acquire such assets from the original providers under the applicable terms.

\newpage
\section*{NeurIPS Paper Checklist}

\begin{enumerate}

\item {\bf Claims}
    \item[] Question: Do the main claims made in the abstract and introduction accurately reflect the paper's contributions and scope?
    \item[] Answer: \answerYes{}
    \item[] Justification: The abstract and Introduction state the data-generation, sandbox, and SFT/RL training contributions, and Secs.~\ref{sec:method}--\ref{sec:experiments} provide the corresponding methodology and empirical evidence. The claims distinguish the four training task families from additional transfer evaluations and comparisons of two open-weight backbones.

\item {\bf Limitations}
    \item[] Question: Does the paper discuss the limitations of the work performed by the authors?
    \item[] Answer: \answerYes{}
    \item[] Justification: Sec.~\ref{sec:conclusion} and Appendix~\ref{sec:supp_limitations} discuss task scope, partial base-mesh sharing for Shape-Key, camera-reframing validity, and model and interaction scaling. Appendices~\ref{sec:supp_imagine} and~\ref{sec:supp_metric_validity} discuss the scope and limitations of the VLM-based evaluations.

\item {\bf Theory assumptions and proofs}
    \item[] Question: For each theoretical result, does the paper provide the full set of assumptions and a complete (and correct) proof?
    \item[] Answer: \answerNA{}
    \item[] Justification: The paper does not present theoretical theorems or proof-based claims. The equations in Sec.~\ref{sec:method} define the task objective, evaluation metrics, and reward computations.

    \item {\bf Experimental result reproducibility}
    \item[] Question: Does the paper fully disclose all the information needed to reproduce the main experimental results of the paper to the extent that it affects the main claims and/or conclusions of the paper (regardless of whether the code and data are provided or not)?
    \item[] Answer: \answerYes{}
    \item[] Justification: The public test set, released model checkpoints, and evaluation code, configurations, and commands provide a path to verify the main benchmark results without retraining. Training data can be generated with the Dataset Pipeline. Secs.~\ref{sec:benchmarks} and~\ref{sec:exp_setup} describe the evaluation protocol and model settings; Appendix~\ref{sec:supp_release_licenses} provides the release entry point and reproduction resources.

\item {\bf Open access to data and code}
    \item[] Question: Does the paper provide open access to the data and code, with sufficient instructions to faithfully reproduce the main experimental results, as described in supplemental material?
    \item[] Answer: \answerYes{}
    \item[] Justification: We provide the test set, model weights, data-generation, training, sandbox, and evaluation method, together with environment setup, evaluation configurations. Training data can be generated with the released pipeline using upstream assets obtained under their respective terms. Appendix~\ref{sec:supp_release_licenses} provides the release entry point and data-access instructions.

\item {\bf Experimental setting/details}
    \item[] Question: Does the paper specify all the training and test details (e.g., data splits, hyperparameters, how they were chosen, type of optimizer) necessary to understand the results?
    \item[] Answer: \answerYes{}
    \item[] Justification: Sec.~\ref{sec:benchmarks} defines the evaluation protocol and metrics, Sec.~\ref{sec:exp_setup} reports model, SFT, RL, and inference settings, and the Appendix provides prompt templates, render settings, and dataset split configurations.

\item {\bf Experiment statistical significance}
    \item[] Question: Does the paper report error bars suitably and correctly defined or other appropriate information about the statistical significance of the experiments?
    \item[] Answer: \answerNo{}
    \item[] Justification: Appendix~\ref{sec:supp_metric_validity} reports 95\% bootstrap confidence intervals and p-values for the VLM--metric correlations, and Appendix~\ref{sec:supp_camera} reports three camera-reframing repeats per task and model. The main model comparisons do not report error bars, confidence intervals, or significance tests.

\item {\bf Experiments compute resources}
    \item[] Question: For each experiment, does the paper provide sufficient information on the computer resources (type of compute workers, memory, time of execution) needed to reproduce the experiments?
    \item[] Answer: \answerNo{}
    \item[] Justification: Sec.~\ref{sec:exp_setup} and Appendix~\ref{sec:supp_compute_resources} report hardware and resource estimates for the original 8B experiments. A complete compute accounting for the additional backbone, transfer, and VLM evaluations is not reported.
    
\item {\bf Code of ethics}
    \item[] Question: Does the research conducted in the paper conform, in every respect, with the NeurIPS Code of Ethics \url{https://neurips.cc/public/EthicsGuidelines}?
    \item[] Answer: \answerYes{}
    \item[] Justification: The work does not involve human subjects or personal data, and the release policy in Appendix~\ref{sec:supp_release_licenses} is designed to respect upstream asset licenses and redistribution restrictions. Appendix~\ref{sec:supp_impact_safeguards} describes intended use, risks, and safeguards for responsible release.

\item {\bf Broader impacts}
    \item[] Question: Does the paper discuss both potential positive societal impacts and negative societal impacts of the work performed?
    \item[] Answer: \answerYes{}
    \item[] Justification: Appendix~\ref{sec:supp_impact_safeguards} discusses positive impacts for 3D content creation, simulation, education, and accessibility, as well as potential misuse in deceptive or infringing 3D content workflows.
    
\item {\bf Safeguards}
    \item[] Question: Does the paper describe safeguards that have been put in place for responsible release of data or models that have a high risk for misuse (e.g., pre-trained language models, image generators, or scraped datasets)?
    \item[] Answer: \answerYes{}
    \item[] Justification: Appendix~\ref{sec:supp_impact_safeguards} discusses intended use and misuse risks, and describes planned safeguards for the dataset and model release, including usage restrictions and exclusion of personal data and restricted upstream raw assets.

\item {\bf Licenses for existing assets}
    \item[] Question: Are the creators or original owners of assets (e.g., code, data, models), used in the paper, properly credited and are the license and terms of use explicitly mentioned and properly respected?
    \item[] Answer: \answerYes{}
    \item[] Justification: Appendix~\ref{sec:supp_release_licenses} lists the upstream assets, models, and software, their licenses or provider terms, and redistribution constraints. It includes both trained backbones and the models used for teacher trajectories, generated goal images, and visual assessment; restricted upstream raw assets are excluded from the planned release.

\item {\bf New assets}
    \item[] Question: Are new assets introduced in the paper well documented and is the documentation provided alongside the assets?
    \item[] Answer: \answerNo{}
    \item[] Justification: The paper documents the dataset formulation, sandbox, training pipeline, and evaluation protocol. The test set, code, model weights, and reproduction instructions are available, and training data can be generated with the Dataset Pipeline. Full dataset and model documentation covering licensing, intended use, and limitations remains planned (Appendix~\ref{sec:supp_release_licenses}).

\item {\bf Crowdsourcing and research with human subjects}
    \item[] Question: For crowdsourcing experiments and research with human subjects, does the paper include the full text of instructions given to participants and screenshots, if applicable, as well as details about compensation (if any)? 
    \item[] Answer: \answerNA{}
    \item[] Justification: The work does not involve crowdsourcing or human-subject experiments. References to human researchers in the paper refer only to the authors' internal curation and predefined asset preparation.

\item {\bf Institutional review board (IRB) approvals or equivalent for research with human subjects}
    \item[] Question: Does the paper describe potential risks incurred by study participants, whether such risks were disclosed to the subjects, and whether Institutional Review Board (IRB) approvals (or an equivalent approval/review based on the requirements of your country or institution) were obtained?
    \item[] Answer: \answerNA{}
    \item[] Justification: No crowdsourcing or human-subject study was conducted, so IRB or equivalent approval is not applicable.

\item {\bf Declaration of LLM usage}
    \item[] Question: Does the paper describe the usage of LLMs if it is an important, original, or non-standard component of the core methods in this research? Note that if the LLM is used only for writing, editing, or formatting purposes and does \emph{not} impact the core methodology, scientific rigor, or originality of the research, declaration is not required.
    \item[] Answer: \answerYes{}
    \item[] Justification: The method and evaluations use Qwen3-14B for shape-key renaming, GPT-5.2 for teacher trajectories, Qwen3-VL-8B and Qwen3.5-9B as editing agents, Gemini for generating Imagine-OOD goal images, and GPT-5.5 for visual preference and metric-validity assessments, as described in Secs.~\ref{sec:data_generation}, \ref{sec:training}, \ref{sec:exp_setup}, and Appendix~\ref{sec:supp_vlm}.

\end{enumerate}

\end{document}